\documentclass[10pt,twocolumn,letterpaper]{article}

\usepackage{cvpr}              %

\newcommand\scalemath[2]{\scalebox{#1}{\mbox{\ensuremath{\displaystyle #2}}}}
\usepackage{amsmath}
\usepackage{amssymb}
\usepackage{mathtools}
\usepackage{amsthm}
\usepackage{multirow}
\usepackage{graphicx}
\usepackage{amsmath}
\usepackage{amssymb}
\usepackage{booktabs}
\usepackage{wrapfig}

\usepackage{float}
\floatstyle{plaintop}
\restylefloat{table}
\usepackage{url} 
\graphicspath{ {figure/} }
\usepackage{tikz}
\usetikzlibrary{shapes,arrows}

\usepackage{amsmath}
\usepackage{amsfonts}
\usepackage{graphicx}
\usepackage{algorithm}
\usepackage{algpseudocode}

\theoremstyle{plain}
\newtheorem{theorem}{Theorem}[section]
\newtheorem{proposition}[theorem]{Proposition}
\newtheorem{lemma}[theorem]{Lemma}

\theoremstyle{definition}

\theoremstyle{remark}

\definecolor{cvprblue}{rgb}{0.21,0.49,0.74}

\def\paperID{12885} %
\def\confName{CVPR}
\def\confYear{2024}

\title{Epistemic Uncertainty Quantification For Pre-trained Neural Networks}

\author{Hanjing Wang\\
Rensselaer Polytechnic Institute\\
110 8th St, Troy, NY, 12180, USA\\
{\tt\small wangh36@rpi.edu}
\and
Qiang Ji\\
Rensselaer Polytechnic Institute\\
110 8th St, Troy, NY, 12180, USA\\
{\tt\small jiq@rpi.edu}
}
\begin{document}
\maketitle

\begin{abstract}
Epistemic uncertainty quantification (UQ) identifies where models lack knowledge. Traditional UQ methods, often based on Bayesian neural networks, are not suitable for pre-trained non-Bayesian models. Our study addresses quantifying epistemic uncertainty for any pre-trained model, which does not need the original training data or model modifications and can ensure broad applicability regardless of network architectures or training techniques. Specifically, we propose a gradient-based approach to assess epistemic uncertainty, analyzing the gradients of outputs relative to model parameters, and thereby indicating necessary model adjustments to accurately represent the inputs. We first explore theoretical guarantees of gradient-based methods for epistemic UQ, questioning the view that this uncertainty is only calculable through differences between multiple models. We further improve gradient-driven UQ by using class-specific weights for integrating gradients and emphasizing distinct contributions from neural network layers. Additionally, we enhance UQ accuracy by combining gradient and perturbation methods to refine the gradients. We evaluate our approach on out-of-distribution detection, uncertainty calibration, and active learning, demonstrating its superiority over current state-of-the-art UQ methods for pre-trained models.

\end{abstract}   
\vspace{-4mm}
\section{Introduction}
\label{sec:intro}

Uncertainty quantification (UQ) is essential in machine learning, especially as models tackle safety-critical tasks like healthcare diagnostics or autonomous navigation. UQ assesses predictive confidence, bolstering model trustworthiness and ensuring safer real-world applications.

There are two types of uncertainty: epistemic and aleatoric. Epistemic uncertainty stems from a lack of knowledge, often due to limited data or model inadequacies, and is potentially reducible given more training data. Aleatoric uncertainty arises from inherent randomness in the data and remains irreducible regardless of data availability. For classification tasks, aleatoric uncertainty in neural networks is often captured by the entropy of the softmax probability distribution. In contrast, Bayesian neural networks (BNNs) provide a systematic framework to estimate epistemic uncertainty by constructing the posterior distribution of model parameters. While the direct calculation of the posterior is often intractable, common methods include Markov Chain Monte Carlo \cite{geman1984stochastic, robert1999monte, tierney1994markov, SGHMC_chen2014, welling2011bayesian, zhang2019cyclical} and variational inference \cite{Maddox_NIPS19_SWAG, louizos2017multiplicative, franchi2020encoding, daxberger2021bayesian, kristiadi2020being} techniques. BNNs then use samples from the posterior distribution to quantify epistemic uncertainty through prediction disagreements.

Our study highlights the importance of measuring epistemic uncertainty with any pre-trained model, avoiding the heavy computational efforts required by BNNs, and focusing on these advantageous conditions:
\begin{itemize}
    \item \textbf{No Extra Knowledge Needed.} Our method works without requiring training or validation datasets, suitable for scenarios with limited data access. 
    \item \textbf{No Model Refinement Required.} We avoid the need for parameter refinement, whether through retraining for new parameters or creating a parameter distribution from the trained model. This minimizes the computational demands and risks associated with model adjustments.
    \item \textbf{Universal Applicability.} UQ is universally applicable for any pre-trained model, free from constraints related to architecture design or specific training techniques.
\end{itemize}

Some existing methods such as gradient-based and perturbation-based UQ can meet the previously stated conditions. Gradient-based UQ is based on the idea that the sensitivity of a model's output to its parameters can indicate prediction uncertainties. Essentially, gradients show the adjustments needed for accurately representing the input, suggesting the model's insufficient knowledge about the input. Perturbation-based UQ, on the other hand, involves applying minor modifications to input data or model parameters to see how these changes impact predictions. If the model is unfamiliar with an input, even small perturbations can significantly change the output. In contrast, if the model knows the input well, it should also handle slight variations. Current gradient-based and perturbation-based methods have several limitations. Firstly, there's a lack of theoretical analysis of their ability to accurately capture epistemic uncertainty. Additionally, their effectiveness often falls short compared to BNNs or even measures like softmax probability entropy \cite{igoe2022useful}. %
Lastly, evaluations of these methods are primarily based on out-of-distribution detection, indicating a need for more comprehensive assessments.

Our study first challenges the idea that epistemic uncertainty is primarily detected through model discrepancies. Specifically, this paper offers theoretical support for general perturbation/gradient-based methods by connecting them to BNNs. We claim that gradient-based and perturbation-based methods are effective for this epistemic UQ with theoretical analysis, suggesting that a single model can recognize their limitations and areas of unfamiliarity.  While conditions for ideal uncertainty estimation may not hold, we introduce three components to enhance gradient-based UQ based on the theoretical insights:
\begin{itemize}
    \item \textbf{Class-specific Gradient Weighting.} In classification tasks, it's possible to calculate gradients for each class. We aim to develop a more effective uncertainty score by combining these gradients using class-specific weights. 
    \item \textbf{Selective Gradient Computation.} We introduce layer-selective gradients, focusing on gradients from deeper layers of the neural network, as they are more representative of epistemic uncertainty.
    \item \textbf{Smoothing Noisy Gradients.} We combine gradients with perturbation techniques, applying slight input perturbations and aggregating gradients for each perturbed input to smooth the raw gradients.
\end{itemize}

\vspace{-2mm}
\section{Related Work}
\label{sec:related_work}
\vspace{-1mm}
\noindent \textbf{Gradient-based UQ.} Gradient-based techniques leverage gradient information for uncertainty estimation. %
The key difference among gradient-based methods is their strategy for deriving an uncertainty score by computing and integrating gradients. For example, Lee et al. \cite{lee2020gradients} backpropagated the cross entropy loss using the network’s predicted probabilities with a confounding label. Huang et al \cite{huang2021importance} computed the gradients of the KL divergence between the predicted distribution and uniform distribution, where epistemic uncertainty is quantified via the norm of gradients. Meanwhile, Igoe et al. \cite{igoe2022useful} introduced ExGrad, which scales the gradients of each class based on its associated predicted probability. While Lee et al. \cite{lee2020gradients} offered intuitive reasoning, gradient-based UQ methods lack solid theoretical foundations, relying primarily on empirical results. As highlighted by Igoe et al. \cite{igoe2022useful}, these gradient-centric UQ methods often don't outperform simple measures like measuring the distance of the predicted distribution to a uniform one, which primarily captures aleatoric uncertainty. This motivates us to propose an improved approach for epistemic UQ, utilizing gradient information more effectively.

\noindent \textbf{Perturbation-based UQ.} Perturbation-based methods \cite{alarab2022adversarial,alarab2023uncertainty,ledda2023dropout,mi2022training} measure epistemic uncertainty by introducing small modifications to the inputs or parameters and observing the resulting variations in the outputs. Alarab et al. \cite{alarab2022adversarial,alarab2023uncertainty} employed adversarial attacks to the inputs to estimate uncertainty. Ledda et al. \cite{ledda2023dropout} introduced extra dropout layers into a trained network that initially did not include dropout, utilizing these layers for epistemic UQ. Despite their empirical effectiveness, these methods lack theoretical support, motivating our theoretical analysis.

\noindent \textbf{Other UQ Methods for Pre-trained Models.} Contrary to perturbation-based and gradient-based techniques, the UQ methods outlined in this section mandate specific prerequisites in pre-trained models for effective epistemic UQ. Several methods are proposed to capture uncertainty through a special design of the network. For example, MC-dropout \cite{gal2016dropout} assumes the network incorporates dropout layers during both training and testing for UQ. Similarly, MC-batchnorm \cite{teye2018bayesian} employs a deterministic network trained with batch normalization, which is also maintained during testing for UQ. Additionally, some methods require extra information beyond the model architecture. Laplace Approximation methods \cite{MacKay1992, daxberger2021bayesian, Ritter_ICLR18_laplace, lee2020estimating} estimate the posterior distribution by approximating it as a Gaussian distribution and performing a Taylor expansion around the learned parameters of a pre-trained model. However, these methods necessitate access to the training data to approximate the posterior accurately.  Utilizing the training data, Schweighofer et al. \cite{schweighofer2023quantification} explored the parameter space around a pre-trained model to identify alternative modes of the posterior distribution.

\vspace{-2mm}
\section{Theoretical Analysis}
In this section, we first introduce the epistemic UQ of a pre-trained model. We then demonstrate how perturbation-based UQ methods can, under certain conditions, serve as effective approximations to BNNs for the estimation of epistemic uncertainty. Finally, we offer a theoretical analysis of the capacity of gradient-based techniques in quantifying epistemic uncertainty, highlighting the connections between perturbation-based and gradient-based approaches.
\vspace{-4mm}
\subsection{Epistemic Uncertainty Quantification}
\textbf{General Notations and Assumptions.} Consider $x$ as the input, $y$ as the target variable, and $\mathcal{D}$ as the training data. We focus on classification tasks. Without loss of generality, we assume the pre-trained model is a single deterministic NN that provides a probability vector $p(y|x,\theta^*) \in [0,1]^C$, where $\theta^*$ are the pre-determined model parameters and $C$ is the class count. We use $\mathbb{E}(\cdot)$ for expectation, $\mathcal{H}(\cdot)$ for entropy, and $\mathcal{I}(\cdot;\cdot)$ to represent mutual information.

\noindent \textbf{Epistemic UQ for BNNs.} BNNs treat the neural network parameters, denoted as $\theta$, as random variables governed by a posterior distribution $p(\theta|\mathcal{D})$. 
A widely used method to estimate epistemic uncertainty involves calculating the mutual information between $y$ and $\theta$, expressed as:
\begin{equation}\label{I_bnn}
    \mathcal{I}(y;\theta|x,\mathcal{D}) = \mathbb{E}_{p(\theta|\mathcal{D})} \left[ \text{KL}(p(y|x,\theta)||p(y|x,\mathcal{D}))\right]
\end{equation}
where KL represents the Kullback–Leibler divergence.

\noindent \textbf{Desired Epistemic UQ for a Pre-trained Model.} While the ideal Bayesian prediction $p(y|x,\mathcal{D})$ remains intractable, Schweighofer et al. \cite{schweighofer2023quantification} approximated it in Eq.~\eqref{I_bnn} using $p(y|x,\theta^*)$ for quantifying epistemic uncertainty of a pre-trained model parameterized by $\theta^*$. They aimed to estimate the prediction difference between $\theta^*$ and samples from the posterior distribution, which is formulated as follows: %
\begin{equation}\label{ideal_ue}
    U_e(x) = \mathbb{E}_{p(\theta|\mathcal{D})} \left[ \text{KL}(p(y|x,\theta)||p(y|x,\theta^*))\right].
\end{equation}
Considering $\theta = \theta^* + \Delta \theta$, we can reformulate $U_e(x)$ from Eq.~\eqref{ideal_ue} by performing perturbations on $\theta^*$:
\begin{equation}
    \label{U_e_perturb}
    \mathbb{E}_{p(\Delta \theta|\mathcal{D})} \left[ \text{KL}(p(y|x,\theta^* + \Delta \theta)||p(y|x,\theta^*))\right].
\end{equation}

\subsection{Perturbation-based UQ}
In this section, we aim to explore the connection between perturbation-based UQ and BNNs for epistemic UQ. Proposition \ref{proposition1} provides a theoretical analysis indicating that, in certain conditions, perturbation-based methods can effectively approximate BNNs.
\vspace{-2mm}
\begin{proposition}\label{proposition1} Assume the model parameters $\theta^*$ are learned given sufficient in-distribution training data $\mathcal{D}$, i.e., $|\mathcal{D}|\rightarrow \infty$. Under mild regularity conditions (i.e., the likelihood function of $\theta$ is continuous, $\theta^*$ is not on the boundary of the parameter space), perturbing $\theta^*$ by adding a small Gaussian noise can accurately approximate the posterior distribution $p(\theta|\mathcal{D})$. For example,
\begin{equation}
    \theta^* + \sigma \epsilon \sim \mathcal{N}(\theta^*,\sigma^2 I) \quad \text{where} \quad \epsilon \sim \mathcal{N}(0,I)
\end{equation}
and 
\begin{equation} \label{convergence}
    \sup_{\theta} |p(\theta|\mathcal{D}) - \mathcal{N}(\theta;\theta^*, \sigma^2 I)| \rightarrow 0 
\end{equation}
where $\sigma \rightarrow 0$ is a small positive constant.
\end{proposition}
\vspace{-1mm}
The proof of Proposition \ref{proposition1} is based on the Bernstein-von Mises theorem \cite{kleijn2012bernstein,gelman2011induction}. This theorem indicates that, given infinite training data, the posterior distribution tends toward a Gaussian distribution. Moreover, as the number of training data increases, this Gaussian distribution narrows. The detailed proof is shown in Appendix A. Proposition \ref{proposition1} underscores the potential of perturbation-based methods to approximate BNNs, especially in cases where the training data is comprehensive and the model is robustly trained. In this scenario, $\mathcal{N}(0,\sigma^2 I)$ closely approximates $p(\Delta \theta|\mathcal{D})$. %
It is worth noting that an adequately large dataset $\mathcal{D}$ in Eq.~\eqref{convergence} makes diagonal covariance sufficient for reliable uncertainty measures. %
The assumption $|\mathcal{D}| \rightarrow \infty$ is significant as they (1) establish conditions for ideal uncertainty estimation and (2) enable the estimation of epistemic uncertainty for any pre-trained model in a challenging scenario, assuming no extra knowledge or learning is needed. While the assumptions are generally hard to achieve, their strict satisfaction isn't necessary when most datasets have sufficient samples and general data augmentation is applied during training. Furthermore, we can derive an upper bound for the total variation distance, $\text{D}_{\text{TV}}(p(\theta|\mathcal{D}), N(\theta;\theta^*,\sigma^2 I)) (\sigma \rightarrow 0)$, to analyze the validity of Proposition \ref{proposition1} for finite $\mathcal{D}$:
\begin{proposition}\label{upper_bound}
Denote $v(\theta) = -\frac{1}{|\mathcal{D}|}\log p(\theta) - \frac{1}{|\mathcal{D}|} \sum_{(x,y) \in \mathcal{D}} \log p(y|x,\theta)$ with $p(\theta)$ as a pre-defined prior distribution. Under Proposition \ref{proposition1} and the regularity constraints on $v(\theta)$ from Sec. 2.2 of \cite{katsevich2023tight}, we have
\begin{equation} \label{eq1}\small
\begin{split}
    \scalemath{1.0}{\text{D}_{\text{TV}}} & \scalemath{1.0}{\leq \sqrt{\frac{1}{2} \text{KL}(\mathcal{N}(\theta^*,\sigma^2 I)||\mathcal{N}(\theta^*,-H(\theta^*)^{-1})}} \\ 
    & \scalemath{1.0}{+ c (c_3^2(v) +c_4(v)) \frac{d^2}{|\mathcal{D}|} + |c_3(v)| \frac{d}{\sqrt{|\mathcal{D}|}}}.
\end{split}
\end{equation}
$H(\theta^*) = \nabla_{\theta^*}^2 \log p(\theta^*|\mathcal{D})$. $d$ is the dimension of $\theta$, $c$ is an absolute constant, and $c_3(v), c_4(v)$ are constants computed from third/fourth-order derivatives of $v$.
\end{proposition} \vspace{-1mm}
Note that $\mathcal{N}(\theta^*,-H(\theta^*)^{-1})$ is the Laplacian approximation of $p(\theta|\mathcal{D})$ and $H(\theta^*)^{-1} \rightarrow \mathbf{0}$ as $|\mathcal{D}|\rightarrow \infty$ based on the Bernstein-von Mises theorem. Hence, the upper bound $\rightarrow$ 0 as $|\mathcal{D}| \rightarrow \infty$. It implies a good convergence when $|\mathcal{D}|\gg d^2$. The proof will be shown in Appendix A. While perturbations in the parameter space have a clear link to BNNs, input-space perturbations offer more understandable modifications. The large parameter space and complex non-linear transformations in NNs make it challenging to identify effective parameter-space perturbations. This complexity also hinders the ability to understand how perturbations operate within neural networks. In contrast, applying perturbations directly to the input is relatively easier, and these perturbations can be visualized.  Intuitively, when a model has limited knowledge about an input (indicative of high epistemic uncertainty), it probably also lacks knowledge about the nearby regions of that input. A slight perturbation to the input might induce a notable change in prediction. Additionally, the following proposition suggests parameter-space and input-space perturbations are transferable under certain conditions with proof shown in Appendix A:
\begin{proposition}\label{proposition2}
    Denote $f(x,\theta)$ as a neural network parameterized by $\theta$ with input $x$. For any small perturbation $\Delta \theta \rightarrow 0$, there exists a small perturbation $\Delta x$ that fulfills 
    \begin{equation}\label{equal_perturbation}
        f(x,\theta + \Delta \theta) =  f(x + \Delta x,\theta).
    \end{equation}
    Conversely, for any small input-space perturbation $\Delta x$, there exists a small perturbation applied on the first layer parameters that satisfies Eq.~\eqref{equal_perturbation}. Based on Eq.~\eqref{equal_perturbation},
    \begin{equation}
    \label{U_e_perturb_x}
    \begin{split}
        U_e(x) &= \mathbb{E}_{\Delta \theta} \left[ \text{\normalfont KL}(p(y|x,\theta^* + \Delta \theta)||p(y|x,\theta^*))\right] \\
        &=\mathbb{E}_{\Delta x} \left[ \text{\normalfont KL}(p(y|x + \Delta x,\theta^*)||p(y|x,\theta^*))\right].
    \end{split}
\end{equation}
\end{proposition}

As a result, under the conditions of Proposition \ref{proposition1} and \ref{proposition2}, suitable perturbations in the input space and subsequent evaluation of output variations can also effectively emulate the behavior of BNNs to measure epistemic uncertainty. Note that when both $\Delta \theta$ and $\Delta x$ are small, Eq.~\eqref{equal_perturbation} can be well-approximated by the first-order Taylor expansion of both sides. This reveals a linear relation between $\Delta x$ and $\Delta \theta$. Thus, for small Gaussian $\Delta \theta$ as per Proposition. \ref{proposition1}, $\Delta x$ is also Gaussian and sampling $\Delta x\sim \mathcal{N}(0,\sigma^2 I)$ with a small $\sigma$ is a suitable approximation.

\subsection{Gradient-based UQ}
Gradients tracing from the model's output back to its parameters are indicative of epistemic uncertainty. Fundamentally, these gradients measure the slight changes necessary in the model's parameters to better represent the input. When faced with an unfamiliar input, the model needs more substantial adjustments in its parameters. Conversely, for familiar, in-distribution samples, these gradients are close to zero. This concept is demonstrated in Proposition \ref{p3}:

\begin{proposition} \label{p3}
Let us assume the neural network $f$ parameterized by $\theta$ has sufficient complexity and it is trained with a sufficiently large dataset $\mathcal{D}$ ($|\mathcal{D}|\rightarrow \infty$) in the neighborhood of an in-distribution input $x$. Under mild regularity conditions (i.e., the likelihood function of $\theta$ is continuous, $\theta^*$ is not on the boundary of the parameter space), the optimized $\theta^*$ achieves global optimality in $x$'s neighborhood, denoted $\mathcal{N}(x)$. For $\forall x +\Delta x \in \mathcal{N}(x)$, we have:
\begin{equation}
    \frac{\partial f(x,\theta^*)}{\partial \theta^*} = 0 \quad \quad \frac{\partial f(x+\Delta x,\theta^*)}{\partial \theta^*} = 0 \quad\quad .
\end{equation}

\end{proposition}

Proposition \ref{p3}, whose proof is in Appendix A, theoretically supports using gradients as epistemic uncertainty indicators, as they may reflect data density and often approach zero for in-distribution data. Intuitively, out-of-distribution data often has non-zero gradients, as it is typically not well-represented by the model, suggesting a need for larger model adjustments to recognize them. Proposition \ref{p3} also indicates that the gradients around the neighborhood of in-distribution $x$ are small. This inspires us to merge the gradients of neighboring samples with the original sample, creating smoother gradients to estimate epistemic uncertainty. 

In exploring the relationship between perturbation-based and gradient-based UQ, it's evident that both approaches fundamentally depend on the sensitivity of the model's outputs to its parameters. While perturbation-based UQ directly modifies parameters or inputs and observes the resulting variations in outputs, gradient-based UQ achieves a similar understanding indirectly through gradient analysis. They do not actively change the parameters but employ gradients directly as a tool for constructing the uncertainty score. The gradients can be viewed as a first-order approximation of the changes observed in perturbation-based methods. Therefore, under small perturbations, both methods are likely to yield similar insights into the model's uncertainty. Building on the connection between perturbation-based and gradient-based methods, the effectiveness of perturbation-based methods by Proposition \ref{proposition1} in measuring epistemic uncertainty suggests a similar potential in gradient-based approaches. The subsequent proposition bridges gradient-based UQ with perturbation-based UQ mathematically:
\begin{proposition} \label{p4}
    The epistemic uncertainty derived by the expected gradient norm can serve as an upper bound compared to the uncertainty produced by perturbation-based methods when the perturbations are small. 
    \begin{equation}\label{upper_bound_gradient}
        \begin{split}
            &\mathbb{E}_{p(\Delta \theta)} \left[ \text{\normalfont KL}(p(y|x,\theta^* )||p(y|x,\theta^* + \Delta \theta))\right] \\
            & \leq \sum_{c=1}^C p(y=c|x,\theta^*)\left \lVert \frac{\partial \log p(y=c|x,\theta^*)}{\partial \theta^*} \right \rVert \mathbb{E}_{p(\Delta \theta)} [||\Delta \theta||] \\
            & \propto \mathbb{E}_{y \sim p(y|x,\theta^*)}\left [ \left \lVert \frac{\partial \log p(y|x,\theta^*)}{\partial \theta^*} \right \rVert \right ] \text{(\normalfont ExGrad \cite{igoe2022useful})}
        \end{split}
    \end{equation}
where $\Delta \theta \rightarrow 0$ and $\mathbb{E}_{p(\Delta \theta)} [||\Delta \theta||]$ is independent of $x$.
\end{proposition}
It is notable that the uncertainty expressed by perturbation-based methods in Eq.~\eqref{upper_bound_gradient} employs the reverse KL divergence, in contrast to Eq.~\eqref{ideal_ue}. Nevertheless, it remains a valid uncertainty metric. Proposition \ref{p4} further implies the necessity to adjust the gradients using the associated probability, indicating different treatments for the gradient of each class. Given the link between gradient-based UQ and perturbation-based techniques, under the conditions of Proposition \ref{proposition1}, the epistemic uncertainty derived from gradient-based methods might also offer a good approximation to the epistemic uncertainty quantified by BNNs. Although the assumptions underlying these propositions might not always hold, it is noteworthy that ExGrad (as detailed in Eq.\eqref{upper_bound_gradient}) could surpass perturbation-based methods, especially in scenarios where perturbations significantly deviate from the optimum. 

\section{Proposed Method}

Our proposed method, rooted in solid theoretical foundations, brings three key advancements to gradient-based UQ. First, as highlighted in Proposition \ref{p4}, it underscores the necessity of assigning distinct weights to the gradients of each class.  The proposed method aims to explore these weights to improve UQ effectiveness. Second, previous approaches uniformly apply the gradient to the parameters across all neural network layers, which may be unsuitable for practical applications. Lastly,  motivated by Proposition \ref{proposition2}, we combine input perturbations with gradients for enhanced performance. Proposition \ref{p3} suggests leveraging gradients from samples near the target input when constructing the uncertainty score. By integrating gradients with perturbations, our approach seeks to enhance UQ by utilizing a smoother gradient representation.

\subsection{Class-specific Gradient Weighting}
When Eq.~\eqref{upper_bound_gradient} suggests weighing the gradient of each class by its associated probability, the method can face challenges. It can produce overconfident results, where a high model probability overshadows the contributions from gradients of other classes, potentially leading to an overemphasis. This might neglect important insights about the model's decision boundaries and uncertain regions indicated by other class gradients. While the gradients of the class with the maximum probability still take the largest weight, we utilize the L2 probability-weighted gradient norm to normalize gradients with a square root:
\begin{equation}\label{rngrad}
    U_{\text{REGrad}}(x) = \sum_{c=1}^C \sqrt{ p(y=c|x,\theta^*)\left \lVert \frac{\partial \log p(y=c|x,\theta^*)}{\partial \theta^*} \right \rVert_2^2}.
\end{equation}
The proposed L2 root-normalized expected gradient method (REGrad) effectively mitigates overconfidence issues, ensuring a more balanced incorporation of gradient information from all classes. 

\subsection{Layer-selective Gradients}

In deep neural networks, each layer captures distinct feature representations, with initial layers extracting low-level features such as edges/texture and deeper layers interpreting complex patterns. Thus, the uncertainty at each layer varies, ranging from ambiguity in basic feature recognition to confusion in decision boundaries. Gradients from deeper layers excel at discerning classification patterns and may struggle to represent out-of-distribution samples, thus being more indicative of epistemic uncertainty.

Recognizing the hierarchical structure of neural networks, we introduce a weighting scheme that assigns importance to layers based on their depth as follows:
\begin{equation}
    \label{layer-selection}
    \left \lVert \frac{\partial \log p(y|x,\theta^*)}{\partial \theta^*} \right \rVert \xrightarrow[\text{selective}]{\text{layer}} \sum_{\theta_l^* \in \theta^*} a_l \left \lVert \frac{\partial \log p(y|x,\theta^*)}{\partial \theta_l^*} \right \rVert .
\end{equation}
Based on Eq.~\eqref{layer-selection}, our approach does not treat all parameter gradients equally. Instead, we apply a coefficient $a_l$ to the gradients of each layer $\theta_l$, indexed by $l$. This coefficient progressively increases from early to deeper layers. Specifically, we use an exponential weighting mechanism, where the weight for layer $l$ is defined as $a_l = \exp(\lambda l)$. Here, $\lambda > 0$ is a hyperparameter controlling the rate of exponential weight increase in deeper layers. By giving more weight to gradients from deeper layers, our method enhances the role of these gradients in measuring uncertainty. This approach emphasizes classification-specific insights and decision-making aspects crucial for detecting out-of-distribution instances. Notably, integrating these layer-selective gradients can improve any gradient-based UQ approach without adding computational complexity.

\begin{table*}[ht]\small
\fontsize{8}{11}\selectfont
	\caption{OOD detection results for AUROC (\%) $\uparrow$ and AUPR (\%)  $\uparrow$ with epistemic uncertainty.  ``*" represents our method (REGrad + layer-selective + perturbation). The experiments are aggregated over three independent runs.} \vspace{-2mm}
	\label{tab:ood_result}
	\centering
\begin{tabular}{|l|cc|cc|cc|cc|cc|l|}
\hline
	\multirow{2}{*}{Method} &\multicolumn{2}{|c|}{MNIST $\rightarrow$ Omniglot} &\multicolumn{2}{|c|}{MNIST $\rightarrow$ FMNIST} & \multicolumn{2}{|c|}{C10$ \rightarrow$ SVHN} & \multicolumn{2}{|c|}{C10$ \rightarrow$ LSUN} &\multirow{2}{*}{Avg}\\ \cline{2-9}
	&AUROC&AUPR &AUROC &AUPR  &AUROC &AUPR &AUROC &AUPR\\
\hline  
\multirow{1}{*}{NEGrad} &41.99 $\pm \scalebox{0.8}{1.75}$&43.49 $\pm \scalebox{0.8}{0.91}$&40.99 $\pm \scalebox{0.8}{3.51}$&44.41 $\pm \scalebox{0.8}{0.91}$&72.80 $\pm \scalebox{0.8}{1.27}$&64.47 $\pm \scalebox{0.8}{1.09}$&66.96 $\pm \scalebox{0.8}{2.05}$&58.51  $\pm \scalebox{0.8}{1.64}$&54.20\\
\multirow{1}{*}{UNGrad} &92.81 $\pm \scalebox{0.8}{0.09}$&90.79 $\pm \scalebox{0.8}{0.33}$&95.41 $\pm \scalebox{0.8}{0.48}$&94.91 $\pm \scalebox{0.8}{0.34}$&38.07 $\pm \scalebox{0.8}{9.79}$&46.86 $\pm \scalebox{0.8}{9.31}$&17.08  $\pm \scalebox{0.8}{8.56}$ &33.52  $\pm \scalebox{0.8}{9.27}$&63.68\\
\multirow{1}{*}{GradNorm} &16.52 $\pm \scalebox{0.8}{1.47}$&35.47 $\pm \scalebox{0.8}{0.61}$&22.71 $\pm \scalebox{0.8}{2.01}$&40.88 $\pm \scalebox{0.8}{0.61}$&10.73 $\pm \scalebox{0.8}{3.09}$&32.04 $\pm \scalebox{0.8}{1.15}$ &10.57  $\pm \scalebox{0.8}{3.93}$ &32.04  $\pm \scalebox{0.8}{0.89}$&25.12\\
\multirow{1}{*}{Exgrad} &97.55 $\pm \scalebox{0.8}{0.05}$&96.99 $\pm \scalebox{0.8}{0.16}$&98.11 $\pm \scalebox{0.8}{0.14}$&97.98 $\pm \scalebox{0.8}{0.16}$&88.87 $\pm \scalebox{0.8}{0.01}$&84.85 $\pm \scalebox{0.8}{1.15}$&88.23  $\pm \scalebox{0.8}{0.11}$&82.26  $\pm \scalebox{0.8}{1.02}$&91.86\\
\multirow{1}{*}{Perturb $x$} &97.16 $\pm \scalebox{0.8}{0.00}$& 95.65 $\pm \scalebox{0.8}{0.16}$& 97.09  $\pm \scalebox{0.8}{0.14}$&95.39  $\pm \scalebox{0.8}{0.31}$&\textbf{92.35} $\pm \scalebox{0.8}{0.22}$&\textbf{91.57} $\pm \scalebox{0.8}{1.32}$&82.89 $\pm \scalebox{0.8}{0.40}$&72.35 $\pm \scalebox{0.8}{0.55}$&90.56\\
\multirow{1}{*}{Perturb $\theta$} &97.64 $\pm \scalebox{0.8}{0.09}$&97.05 $\pm \scalebox{0.8}{0.24}$&97.99  $\pm \scalebox{0.8}{0.16}$&97.64  $\pm \scalebox{0.8}{0.22}$&89.31 $\pm \scalebox{0.8}{1.26}$&85.80 $\pm \scalebox{0.8}{3.96}$&88.33 $\pm \scalebox{0.8}{0.11}$&82.99 $\pm \scalebox{0.8}{0.27}$&92.09\\
\multirow{1}{*}{MC-AA} &91.23 $\pm \scalebox{0.8}{0.35}$&89.40 $\pm \scalebox{0.8}{0.28}$&95.28  $\pm \scalebox{0.8}{0.44}$&95.24  $\pm \scalebox{0.8}{0.19}$&87.56 $\pm \scalebox{0.8}{0.57}$&82.22 $\pm \scalebox{0.8}{2.08}$&82.96 $\pm \scalebox{0.8}{0.84}$&71.79 $\pm \scalebox{0.8}{1.62}$&86.96\\
\multirow{1}{*}{Inserted Dropout} &97.52 $\pm \scalebox{0.8}{0.14}$&96.88 $\pm \scalebox{0.8}{0.35}$&97.00 $\pm \scalebox{0.8}{0.68}$&95.75 $\pm \scalebox{0.8}{0.79}$&88.84 $\pm \scalebox{0.8}{1.02}$&88.01 $\pm \scalebox{0.8}{1.19}$&88.18 $\pm \scalebox{0.8}{0.62}$&86.93 $\pm \scalebox{0.8}{0.73}$&92.39\\
\multirow{1}{*}{Entropy} &97.70 $\pm \scalebox{0.8}{0.08}$&97.38 $\pm \scalebox{0.8}{0.19}$&98.04 $\pm \scalebox{0.8}{0.22}$&97.94 $\pm \scalebox{0.8}{0.17}$&88.86 $\pm \scalebox{0.8}{0.34}$&84.10 $\pm \scalebox{0.8}{0.36}$&89.84 $\pm \scalebox{0.8}{0.30}$&87.72 $\pm \scalebox{0.8}{0.57}$&92.70\\
\multirow{1}{*}{ExGrad V Term} &97.70 $\pm \scalebox{0.8}{0.02}$&97.39 $\pm \scalebox{0.8}{0.13}$&98.04 $\pm \scalebox{0.8}{0.14}$&97.95 $\pm \scalebox{0.8}{0.26}$&89.04 $\pm \scalebox{0.8}{0.43}$&84.64 $\pm \scalebox{0.8}{0.41}$&89.96 $\pm \scalebox{0.8}{0.18}$&88.09 $\pm \scalebox{0.8}{0.29}$&92.85\\
\multirow{1}{*}{LA} &97.92 $\pm \scalebox{0.8}{0.14}$&97.45 $\pm \scalebox{0.8}{0.29}$&97.95 $\pm \scalebox{0.8}{0.27}$&97.37  $\pm \scalebox{0.8}{0.43}$&91.86 $\pm \scalebox{0.8}{0.27}$ &88.06 $\pm \scalebox{0.8}{0.27}$&86.81 $\pm \scalebox{0.8}{0.27}$&85.04 $\pm \scalebox{0.8}{0.27}$&92.81\\
\multirow{1}{*}{REGrad*} &\textbf{98.19} $\pm \scalebox{0.8}{0.02}$&\textbf{97.95} $\pm \scalebox{0.8}{0.06}$&\textbf{98.78} $\pm \scalebox{0.8}{0.14}$&\textbf{98.79} $\pm \scalebox{0.8}{0.07}$&92.32 $\pm \scalebox{0.8}{0.59}$&89.42 $\pm \scalebox{0.8}{2.81}$&\textbf{91.11}  $\pm \scalebox{0.8}{0.44}$&\textbf{89.93}  $\pm \scalebox{0.8}{0.69}$&\textbf{94.56}\\ 
\hline
\end{tabular}

\vspace{+1mm}
\begin{tabular}{|l|cc|cc|cc|cc|cc|l|}
\hline
	\multirow{2}{*}{Method} &\multicolumn{2}{|c|}{SVHN $\rightarrow$ C10} &\multicolumn{2}{|c|}{SVHN $\rightarrow$ C100} & \multicolumn{2}{|c|}{C100$ \rightarrow$ SVHN} & \multicolumn{2}{|c|}{C100$ \rightarrow$ LSUN} &\multirow{2}{*}{Avg}\\ \cline{2-9}
	&AUROC&AUPR &AUROC &AUPR  &AUROC &AUPR &AUROC &AUPR\\
\hline  
\multirow{1}{*}{NEGrad} &30.26 $\pm \scalebox{0.8}{1.80}$&39.13 $\pm \scalebox{0.8}{1.31}$&30.93 $\pm \scalebox{0.8}{1.24}$&39.38 $\pm \scalebox{0.8}{0.98}$&44.62 $\pm \scalebox{0.8}{2.65}$&42.52 $\pm \scalebox{0.8}{1.96}$&43.99 $\pm \scalebox{0.8}{2.21}$& 42.74 $\pm \scalebox{0.8}{3.28}$ &39.20\\
\multirow{1}{*}{UNGrad} &68.28 $\pm \scalebox{0.8}{3.45}$&69.21 $\pm \scalebox{0.8}{3.18}$& 68.34 $\pm \scalebox{0.8}{2.77}$&68.95 $\pm \scalebox{0.8}{2.42}$& 44.34 $\pm \scalebox{0.8}{7.36}$&43.44 $\pm \scalebox{0.8}{6.66}$&47.46 $\pm \scalebox{0.8}{4.63}$&45.41 $\pm \scalebox{0.8}{5.36}$&56.93\\
\multirow{1}{*}{GradNorm} &24.49 $\pm \scalebox{0.8}{3.75}$& 36.56 $\pm \scalebox{0.8}{2.06}$&25.62 $\pm \scalebox{0.8}{2.72}$ &37.26 $\pm \scalebox{0.8}{1.42}$ &10.67 $\pm \scalebox{0.8}{2.08}$&32.11 $\pm \scalebox{0.8}{0.44}$& 27.53 $\pm \scalebox{0.8}{3.91}$&36.97 $\pm \scalebox{0.8}{2.19}$&28.90\\
\multirow{1}{*}{Exgrad} &88.57 $\pm \scalebox{0.8}{0.61}$& 85.55 $\pm \scalebox{0.8}{1.33}$& 88.25 $\pm \scalebox{0.8}{0.62}$& 85.21 $\pm \scalebox{0.8}{1.14}$& 79.99 $\pm \scalebox{0.8}{1.37}$& 72.68 $\pm \scalebox{0.8}{0.82}$& 76.93 $\pm \scalebox{0.8}{1.64}$& 71.00 $\pm \scalebox{0.8}{2.62}$&81.02\\
\multirow{1}{*}{Perturb $x$} &72.76 $\pm \scalebox{0.8}{0.36}$&60.60 $\pm \scalebox{0.8}{0.26}$& 73.96 $\pm \scalebox{0.8}{0.42}$& 62.27 $\pm \scalebox{0.8}{0.25}$&86.49 $\pm \scalebox{0.8}{1.62}$&86.76 $\pm \scalebox{0.8}{1.72}$& 41.10 $\pm \scalebox{0.8}{1.42}$& 44.35 $\pm \scalebox{0.8}{0.72}$&66.04\\
\multirow{1}{*}{Perturb $\theta$} &89.01 $\pm \scalebox{0.8}{0.22}$& 86.04 $\pm \scalebox{0.8}{0.41}$&88.75 $\pm \scalebox{0.8}{0.14}$&85.83 $\pm \scalebox{0.8}{0.21}$&50.99 $\pm \scalebox{0.8}{2.23}$&51.28 $\pm \scalebox{0.8}{3.08}$&39.42 $\pm \scalebox{0.8}{0.70}$& 42.57 $\pm \scalebox{0.8}{1.07}$&66.74\\
\multirow{1}{*}{MC-AA} &71.53 $\pm \scalebox{0.8}{1.32}$&60.98 $\pm \scalebox{0.8}{1.67}$&72.75 $\pm \scalebox{0.8}{2.12}$&62.55 $\pm \scalebox{0.8}{2.72}$&54.74 $\pm \scalebox{0.8}{1.08}$&59.26 $\pm \scalebox{0.8}{2.49}$&31.61 $\pm \scalebox{0.8}{1.81}$&39.40 $\pm \scalebox{0.8}{2.22}$&56.60\\
\multirow{1}{*}{Inserted Dropout} &88.47 $\pm \scalebox{0.8}{1.23}$&83.97 $\pm \scalebox{0.8}{0.71}$&86.96 $\pm \scalebox{0.8}{1.09}$&82.40 $\pm \scalebox{0.8}{0.88}$&57.39 $\pm \scalebox{0.8}{1.46}$&55.49 $\pm \scalebox{0.8}{2.02}$&59.41 $\pm \scalebox{0.8}{1.61}$&58.30 $\pm \scalebox{0.8}{0.92}$&71.55\\
\multirow{1}{*}{Entropy} &89.64 $\pm \scalebox{0.8}{0.04}$&87.76 $\pm \scalebox{0.8}{2.19}$&89.09 $\pm \scalebox{0.8}{0.13}$&87.25 $\pm \scalebox{0.8}{0.06}$& 53.38 $\pm \scalebox{0.8}{0.49}$&55.20 $\pm \scalebox{0.8}{0.18}$ &42.28 $\pm \scalebox{0.8}{0.99}$&46.07 $\pm \scalebox{0.8}{0.39}$&68.83\\
\multirow{1}{*}{ExGrad V Term} &89.72 $\pm \scalebox{0.8}{0.08}$&87.90 $\pm \scalebox{0.8}{1.87}$&89.19 $\pm \scalebox{0.8}{0.07}$&87.40 $\pm \scalebox{0.8}{0.02}$&50.35 $\pm \scalebox{0.8}{0.42}$&52.66 $\pm \scalebox{0.8}{0.73}$&41.67 $\pm \scalebox{0.8}{1.51}$&45.91 $\pm \scalebox{0.8}{1.12}$&68.10\\
\multirow{1}{*}{LA} &80.73 $\pm \scalebox{0.8}{0.73}$&87.46 $\pm \scalebox{0.8}{1.00}$&88.77 $\pm \scalebox{0.8}{0.78}$&86.06 $\pm \scalebox{0.8}{1.12}$  &80.09 $\pm \scalebox{0.8}{0.12}$&73.63 $\pm \scalebox{0.8}{0.39}$&76.50 $\pm \scalebox{0.8}{0.40}$&71.92 $\pm \scalebox{0.8}{0.48}$&80.65\\
\multirow{1}{*}{REGrad*} &\textbf{91.11} $\pm \scalebox{0.8}{0.19}$ &\textbf{89.63} $\pm \scalebox{0.8}{0.37}$&\textbf{90.37} $\pm \scalebox{0.8}{0.26}$ &\textbf{88.83} $\pm \scalebox{0.8}{0.37}$&\textbf{88.06} $\pm \scalebox{0.8}{1.29}$&\textbf{85.74} $\pm \scalebox{0.8}{1.45}$ &\textbf{79.11} $\pm \scalebox{0.8}{0.32}$&\textbf{74.39} $\pm \scalebox{0.8}{0.67}$&\textbf{85.91}\\ 
\hline
\end{tabular}\vspace{-5mm}
\end{table*}

\subsection{Gradient-Perturbation Integration}

While gradients and perturbations are both valuable for understanding model behavior and uncertainty, we aim to combine their strengths for UQ. Our method calculates gradients for both the original and perturbed inputs to produce a smoother and more informative gradient representation, addressing the issue of noisy raw gradients. As Proposition \ref{p3} suggests, gradients from perturbed inputs offer additional uncertainty insights. This concept aligns with Smilkov et al.'s findings \cite{smilkov2017smoothgrad}, where averaging gradients from perturbed inputs results in smoother gradients. While their research focused on using these gradients for clearer saliency maps in model explanation, we specifically apply these refined gradients to improve UQ. Specifically, for an input $x^0$, we introduce input perturbations $\Delta x \sim \mathcal{N}(0,\sigma^2 I)$, controlled by a hyperparameter $\sigma$. This process generates perturbed inputs $x^1, x^2, \cdots, x^N$. The smoothed gradients are then calculated as follows:
\begin{equation}
    \label{smooth_grad}
    \left \lVert \frac{\partial \log p(y|x_0,\theta^*)}{\partial \theta^*} \right \rVert \xrightarrow[\text{smoothed}]{\text{perturb}}  \left \lVert \frac{1}{N+1}\sum_{i=0}^N \frac{ \partial \log p(y|x_i,\theta^*)}{\partial \theta^*} \right \rVert 
\end{equation}

Eq.~\eqref{smooth_grad} effectively mitigates sharp, uncharacteristic spikes in the gradient space, ensuring a more stable uncertainty measure. It is crucial to note that we maintain a small value for $\sigma$, aligning with the guidelines of Proposition \ref{proposition1}. Calculating smoothed gradients does not complicate the backpropagation process, as it requires only one backward pass for the averaged probability vector. The additional computation mainly involves forward passes for the perturbed inputs. However, with all perturbed inputs prepared, these forward passes can be executed concurrently, leveraging parallel processing for efficiency.

\section{Experiments}
\noindent \textbf{Dataset.} We evaluated our method using benchmark image classification datasets, including MNIST \cite{deng2012mnist}, SVHN \cite{netzer2011reading}, CIFAR-10 (C10) \cite{krizhevsky2014cifar}, and CIFAR-100 (C100) \cite{krizhevsky2009learning}.

\noindent \textbf{Implementation Details.} For MNIST and SVHN, we employed standard CNN architectures. For C10 and C100, we used ResNet18 and ResNet152, respectively. These models were trained following standard protocols to develop deterministic classification models, which then served as the pre-trained models for our epistemic UQ experiments. Detailed experiment settings, baseline implementations, and hyperparameter information are provided in Appendix B. 

\noindent \textbf{Baselines.} Our method (REGrad + layer-selective + perturbation) is compared against various baselines:
\begin{itemize}
    \item Gradient-based Methods:\\
    NEGrad \cite{igoe2022useful} ($||\mathbb{E}_{y \sim p(y|x,\theta^*)}[\nabla_{\theta^*}\log p(y|x,\theta^*)]||$), \\UNGrad \cite{igoe2022useful} ($\mathbb{E}_{y\sim\text{uniform}}[||\nabla_{\theta^*}\log p(y|x,\theta^*)||]$),\\ GradNorm  \cite{huang2021importance} ($||\nabla_{\theta^*} \mathbb{E}_{y\sim\text{uniform}}[\log p(y|x,\theta^*)]||$), \\
    ExGrad \cite{igoe2022useful} ($\mathbb{E}_{y\sim p(y|x,\theta^*)}[||\nabla_{\theta^*}\log p(y|x,\theta^*)]||]$ ).
    \item Perturbation-based Methods: \\Perturb $x$ ($\mathbb{E}_{\Delta x \sim \mathcal{N}(0,\sigma^2 I)} \left[ \text{KL}(p(y|x + \Delta x ,\theta^* )||p(y|x,\theta^*))\right]$), Perturb $\theta$ ($\mathbb{E}_{\Delta \theta \sim \mathcal{N}(0,\sigma^2 I)} \left[\text{KL}(p(y|x,\theta^* + \Delta \theta)||p(y|x,\theta^*))\right]$), MC-AA \cite{alarab2022adversarial}, Inserted Dropout \cite{ledda2023dropout}.
    \item Entropy-based Methods: \\
    Entropy (-$\mathbb{E}_{y \sim p(y|x,\theta^*)}[\log p(y|x,\theta^*)]$),\\
    ExGrad V Term ($\sum_{c} |p(y=c|x,\theta^*)-\frac{1}{C}|$)
    \item Approximated BNN using training data: LA \cite{kristiadi2020being}.
\end{itemize}

\noindent \textbf{Evaluation Tasks.} Section \ref{OOD} presents our evaluation of epistemic UQ performance in out-of-distribution (OOD) detection. Section \ref{calibration} details our uncertainty calibration evaluations. In Section \ref{active_learning}, we discuss our findings from uncertainty-guided active learning evaluations.

\subsection{Out-of-distribution Detection} \label{OOD}

OOD detection is a crucial application for UQ \cite{Malinin_prior_network_NIPS18, malinin2019ensemble}. It aims to identify data points that deviate from the training data distribution by utilizing uncertainty measures. Epistemic uncertainty, inversely correlating with data density, tends to be higher when the model encounters anomalous data.

\noindent \textbf{Experiment Settings.} ~~ For MNIST dataset, OOD samples come from the Omniglot \cite{lake2015human} and Fashion-MNIST (FMNIST) \cite{xiao2017fashion} datasets. In the case of C10 and C100, the OOD datasets are SVHN \cite{netzer2011reading} and LSUN \cite{journals/corr/YuZSSX15}. Conversely, for SVHN as in-distribution data, C10 and C100 serve as OOD datasets. We evaluated OOD detection using two metrics: area under the receiver operating characteristic curve (AUROC) and area under the precision-recall curve (AUPR).

\begin{table*}[ht]\small
\fontsize{8}{11}\selectfont
	\caption{Uncertainty calibration results for MNIST, C10, SVHN, and C100 datasets using rAULC $\uparrow$. ``*" represents our method and the experiments are aggregated over three independent runs.} \vspace{-2mm}
	\label{tab:calibration}
	\centering
\begin{tabular}{|l|c|c|c|c|l|}
\hline
	\multirow{2}{*}{Method} & \multicolumn{1}{|c|}{MNIST} &\multicolumn{1}{|c|}{C10} & \multicolumn{1}{|c|}{SVHN} &\multicolumn{1}{|c|}{C100} &\multirow{2}{*}{Avg}\\ \cline{2-5}
	&rAULC &rAULC&rAULC&rAULC\\
\hline  
\multirow{1}{*}{NEGrad} &0.143 $\pm \scalebox{0.75}{.055}$&0.598 $\pm \scalebox{0.75}{.039}$&-0.861 $\pm \scalebox{0.75}{.042}$&0.127 $\pm \scalebox{0.75}{.050}$ &0.002\\
\multirow{1}{*}{UNGrad}  &0.934 $\pm \scalebox{0.75}{.004}$&0.840 $\pm \scalebox{0.75}{.022}$&0.429 $\pm \scalebox{0.75}{.061}$&0.403 $\pm \scalebox{0.75}{.038}$&0.652\\
\multirow{1}{*}{GradNorm}  &-2.810 $\pm \scalebox{0.75}{.022}$&-1.275 $\pm \scalebox{0.75}{.095}$&-1.176 $\pm \scalebox{0.75}{.026}$&-0.995 $\pm \scalebox{0.75}{.098}$&-1.564\\
\multirow{1}{*}{ExGrad}  &\textbf{0.985} $\pm \scalebox{0.75}{.001}$&0.893 $\pm \scalebox{0.75}{.016}$&0.868 $\pm \scalebox{0.75}{.003}$&0.841 $\pm \scalebox{0.75}{.017}$&0.897\\
\multirow{1}{*}{Perturb $x$}  &\textbf{0.985} $\pm \scalebox{0.75}{.001}$&0.890 $\pm \scalebox{0.75}{.014}$&0.830 $\pm \scalebox{0.75}{.001}$&0.772 $\pm \scalebox{0.75}{.032}$&0.869\\
\multirow{1}{*}{Perturb $\theta$} &0.984 $\pm \scalebox{0.75}{.001}$&0.892 $\pm \scalebox{0.75}{.015}$&0.869 $\pm \scalebox{0.75}{.001}$&0.705 $\pm \scalebox{0.75}{.002}$&0.863\\
\multirow{1}{*}{MC-AA}  &0.969 $\pm \scalebox{0.75}{.002}$&0.879 $\pm \scalebox{0.75}{.018}$&0.807 $\pm \scalebox{0.75}{.000}$&0.784 $\pm \scalebox{0.75}{.017}$&0.860\\
\multirow{1}{*}{Inserted Dropout}&0.973 $\pm \scalebox{0.75}{.003}$&0.891 $\pm \scalebox{0.75}{.005}$&0.851 $\pm \scalebox{0.75}{.006}$&0.783 $\pm \scalebox{0.75}{.003}$&0.874\\
\multirow{1}{*}{Entropy}  &\textbf{0.985} $\pm \scalebox{0.75}{.001}$&0.893 $\pm \scalebox{0.75}{.015}$&0.867 $\pm \scalebox{0.75}{.004}$&\textbf{0.863} $\pm \scalebox{0.75}{.017}$&0.902\\
\multirow{1}{*}{ExGrad V Term}  &\textbf{0.985} $\pm \scalebox{0.75}{.001}$&0.893 $\pm \scalebox{0.75}{.015}$&0.865 $\pm \scalebox{0.75}{.004}$&0.854 $\pm \scalebox{0.75}{.020}$&0.899\\
\multirow{1}{*}{LA} &0.983 $\pm \scalebox{0.75}{.002}$&0.865 $\pm \scalebox{0.75}{.003}$&0.857 $\pm \scalebox{0.75}{.002}$&0.792 $\pm \scalebox{0.75}{.007}$&0.874\\
\multirow{1}{*}{REGrad*} &\textbf{0.985} $\pm \scalebox{0.75}{.001}$&\textbf{0.898} $\pm \scalebox{0.75}{.016}$&\textbf{0.873} $\pm \scalebox{0.75}{.002}$&0.858 $\pm \scalebox{0.75}{.013}$&\textbf{0.904}\\
\hline
\end{tabular}\\
\vspace{-3mm}
\end{table*}

\noindent \textbf{Experiment Analysis} ~~In the OOD detection experiments shown in Table \ref{tab:ood_result}, REGrad demonstrates superior performance over various baselines, including gradient-based, perturbation-based, and entropy-based methods. It achieves an impressive average improvement of 30\% to 60\% over NEGrad, UNGrad, and GradNorm. Moreover, REGrad* consistently outperforms the strongest competitor in each category. Against ExGrad, the state-of-the-art gradient-based method, REGrad* shows average enhancements of 2.7\% and 4.84\% across two tables. Within the perturbation-based category, it notably surpasses Perturb $x$ and Perturb $\theta$ by 19.82\% and 19.12\% for SVHN and C100 datasets, respectively. Compared to entropy-based methods, which mainly capture aleatoric uncertainty, REGrad*'s improvement over Entropy and ExGrad V Term ranged from 1.71\% to 17.76\% in averaged results across both tables. Even against LA, which approximates the posterior distribution of parameters using training data, REGrad* shows considerable improvements. Notably, given its assumption of a single Gaussian posterior and the computational constraints requiring a diagonal or block-diagonal covariance matrix, LA may not always be the most effective. These findings underscore REGrad*'s robustness and advanced capability in epistemic UQ across diverse scenarios.

\subsection{Uncertainty Calibration} \label{calibration}

In this section, we demonstrate the effectiveness of our methods in uncertainty calibration. Uncertainty calibration measures the alignment of a model's predicted uncertainty with its actual performance. Well-calibrated uncertainty is crucial for informed decision-making, as it indicates not just what a model predicts, but also the reliability of those predictions. Traditional calibration metrics, such as the expected calibration error and negative log-likelihood, are unsuitable for our scenario. These metrics typically assess $p(y|x,\theta^*)$, but in our case, $p(y|x,\theta^*)$ remains constant across all UQ baselines when using a pre-trained model $\theta^*$. Therefore, we opt for the relative area under the lift curve (rAULC) as our evaluation metric. Introduced in \cite{postels2021practicality}, rAULC is derived by ordering predictions by increasing uncertainty, then plotting the performance for samples with an uncertainty value below a certain quantile of the uncertainty against that quantile.

\noindent \textbf{Experiment Analysis} The uncertainty calibration results in Table \ref{tab:calibration} show that REGrad* consistently outperforms perturbation-based methods and LA, achieving approximately 3\% to 4\% higher performance. Moreover, our method significantly surpasses most gradient-based methods, such as NEGrad, UNGrad, and GradNorm. However, it exhibits only a marginal improvement over ExGrad. ExGrad's effectiveness in uncertainty calibration may be due to its use of the predictive probability vector for weighting gradients of each class. This suggests that the probability vector could be a reliable indicator of model performance. Additionally, REGrad* shows only a slight advantage over Entropy and ExGrad V Term. This can be attributed to the fact that the well-measured aleatoric uncertainty can also indicate model performance, as noted in \cite{postels2021practicality}. Overall, REGrad* serves as the most effective performance indicator among all evaluated baselines.

\begin{table*}[ht]\small
\fontsize{7.8}{11}\selectfont
	\caption{Active learning results of averaged ACC $\uparrow$ and NLL $\downarrow$ across 10 acquisition cycles for all datasets . ``*" represents our method.} \vspace{-2mm}
	\label{tab:active_learning}
	\centering
\begin{tabular}{|l|cc|cc|cc|cc|cc|}
\hline
	\multirow{2}{*}{Method} & \multicolumn{2}{|c|}{MNIST} &\multicolumn{2}{|c|}{C10} & \multicolumn{2}{|c|}{SVHN} &\multicolumn{2}{|c|}{C100}&\multicolumn{2}{|c|}{Avg}\\ \cline{2-11}
	&ACC &NLL &ACC &NLL &ACC &NLL &ACC &NLL&ACC &NLL \\
\hline  
\multirow{1}{*}{NEGrad} &70.22 $\pm \scalebox{0.75}{3.02}$&1.833 $\pm \scalebox{0.75}{.04}$&39.13 $\pm \scalebox{0.75}{.72}$ &2.074 $\pm \scalebox{0.75}{.02}$ &65.12 $\pm \scalebox{0.75}{2.25}$ &1.827 $\pm \scalebox{0.75}{.02}$ &13.16 $\pm \scalebox{0.75}{.27}$ &4.507 $\pm \scalebox{0.75}{.01}$ &46.91&2.560\\
\multirow{1}{*}{UNGrad}  &69.56 $\pm \scalebox{0.75}{2.94}$&1.867 $\pm \scalebox{0.75}{.03}$ &
  38.90 $\pm \scalebox{0.75}{.74}$& 2.071 $\pm \scalebox{0.75}{.01}$& 67.27 $\pm \scalebox{0.75}{2.17}$& 1.808 $\pm \scalebox{0.75}{.02}$ &12.56 $\pm \scalebox{0.75}{.58}$&4.508 $\pm \scalebox{0.75}{.01}$&47.07&2.564\\
\multirow{1}{*}{GradNorm}  &66.85 $\pm \scalebox{0.75}{3.18}$&1.884 $\pm \scalebox{0.75}{.04}$& 39.02 $\pm \scalebox{0.75}{.81}$ &2.069 $\pm \scalebox{0.75}{.01}$&65.43 $\pm \scalebox{0.75}{2.30}$ & 1.827 $\pm \scalebox{0.75}{.03}$&12.89 $\pm \scalebox{0.75}{.33}$&4.509 $\pm \scalebox{0.75}{.01}$&46.05&2.572\\
\multirow{1}{*}{ExGrad }  &69.91 $\pm \scalebox{0.75}{2.21}$&1.847 $\pm \scalebox{0.75}{.03}$&  39.20 $\pm \scalebox{0.75}{.87}$& 2.069$\pm \scalebox{0.75}{.01}$ &66.76 $\pm \scalebox{0.75}{2.11}$& 1.812 $\pm \scalebox{0.75}{.02}$&12.55 $\pm \scalebox{0.75}{.32}$&4.506 $\pm \scalebox{0.75}{.00}$&47.11&2.559\\
\multirow{1}{*}{Perturb $x$}  &71.10 $\pm \scalebox{0.75}{3.43}$&1.844 $\pm \scalebox{0.75}{.05}$&38.57 $ \pm \scalebox{0.75}{.72}$& 2.072$ \pm \scalebox{0.75}{.01}$&64.00 $\pm \scalebox{0.75}{2.39}$& 1.838 $\pm \scalebox{0.75}{.02}$&12.23 $\pm \scalebox{0.75}{.27}$&4.511 $\pm \scalebox{0.75}{.00}$&46.47&2.566\\
\multirow{1}{*}{Perturb $\theta$}  &71.44 $\pm \scalebox{0.75}{2.43}$&1.826 $\pm \scalebox{0.75}{.03}$& 38.97 $\pm \scalebox{0.75}{1.03}$ & 2.070 $\pm \scalebox{0.75}{.02}$&66.31 $ \pm \scalebox{0.75}{2.62}$&  1.817 $\pm \scalebox{0.75}{.03}$&13.17 $\pm \scalebox{0.75}{.37}$&4.506 $\pm \scalebox{0.75}{.01}$&47.47&2.555\\
\multirow{1}{*}{MC-AA}  &71.69 $\pm \scalebox{0.75}{.42}$&1.809 $\pm \scalebox{0.75}{.00}$&38.57 $\pm \scalebox{0.75}{.72}$& 2.073 $\pm \scalebox{0.75}{.01}$&  64.56 $\pm \scalebox{0.75}{1.17}$& 1.831 $ \pm \scalebox{0.75}{.02}$&12.14 $\pm \scalebox{0.75}{.40}$&4.515 $\pm \scalebox{0.75}{.01}$&46.74&2.557\\
\multirow{1}{*}{Inserted Dropout}&72.84 $\pm \scalebox{0.75}{0.62}$&1.836 $\pm \scalebox{0.75}{.00}$&38.34 $\pm \scalebox{0.75}{.11}$&2.075 $\pm \scalebox{0.75}{.01}$&66.99 $\pm \scalebox{0.75}{2.58}$&1.811 $\pm \scalebox{0.75}{.02}$&12.59 $\pm \scalebox{0.75}{.13}$&4.507 $\pm \scalebox{0.75}{.00}$&47.69&2.557\\
\multirow{1}{*}{Entropy}  &70.40 $\pm \scalebox{0.75}{4.33}$&1.846 $\pm \scalebox{0.75}{.06}$& 38.73 $\pm \scalebox{0.75}{.65}$& 2.074 $\pm \scalebox{0.75}{.01}$&63.42 $\pm \scalebox{0.75}{2.86}$& 1.845 $\pm \scalebox{0.75}{.03}$&11.57 $\pm \scalebox{0.75}{.41}$&4.521 $\pm \scalebox{0.75}{.01}$&46.03&2.572\\
\multirow{1}{*}{ExGrad V Term}  &68.53 $\pm \scalebox{0.75}{4.40}$& 1.881 $\pm \scalebox{0.75}{.04}$&  38.66 $\pm \scalebox{0.75}{.67}$&
 2.071 $\pm \scalebox{0.75}{.01}$&62.32 $\pm \scalebox{0.75}{2.87}$& 1.856 $\pm \scalebox{0.75}{.03}$&11.30 $\pm \scalebox{0.75}{.38}$&4.520 $\pm \scalebox{0.75}{.01}$&45.20&2.582\\
\multirow{1}{*}{LA}  &62.65 $\pm \scalebox{0.75}{.58}$&1.935 $\pm \scalebox{0.75}{.02}$&37.56 $\pm \scalebox{0.75}{.29}$&2.083 $\pm \scalebox{0.75}{.00}$&63.44 $\pm \scalebox{0.75}{.75}$ & 1.846 $\pm \scalebox{0.75}{.00}$&12.59 $\pm \scalebox{0.75}{.12}$&\textbf{4.505} $\pm \scalebox{0.75}{.00}$&44.06&2.592\\
\multirow{1}{*}{REGrad*} &\textbf{75.31} $\pm \scalebox{0.75}{3.48}$&\textbf{1.808} $\pm \scalebox{0.75}{.03}$& \textbf{39.28} $\pm \scalebox{0.75}{.83}$ & \textbf{2.067} $\pm \scalebox{0.75}{.01}$ & \textbf{67.37} $\pm \scalebox{0.75}{2.13}$ & \textbf{1.807} $\pm \scalebox{0.75}{.02}$ &\textbf{13.37} $\pm \scalebox{0.75}{.25}$&\textbf{4.505} $\pm \scalebox{0.75}{.00}$&\textbf{48.83}&\textbf{2.547}\\
\hline
\end{tabular}\\
\vspace*{-3mm}
\end{table*}

\subsection{Active Learning} \label{active_learning}
Epistemic uncertainty is important in active learning (AL), where the aim is to annotate data points with the highest uncertainty for retraining the model, thus targeting areas where the model has minimal information. Initially, we train the model with $m_1$ data points. Over 10 acquisition cycles, we select $m_2$ new data points at each cycle from the unused training pool, focusing on those with the highest epistemic uncertainty to retrain the model along with the previously chosen data. Our experiments include MNIST, C10, SVHN, and C100 datasets, setting $m_1$ at 20, 500, 500, and 1000, and $m_2$ at 20, 100, 100, and 500 for each dataset, respectively. Due to the experiments' reliance on smaller data subsets, we use a standard CNN with two convolutional and two fully connected layers across all datasets. Implementation details are available in Appendix B. We evaluate the model's performance by measuring accuracy (ACC) and negative log-likelihood (NLL) on the original testing data. 

\noindent \textbf{Experiment Results.} Table \ref{tab:active_learning} presents the average ACC and NLL across 10 active learning acquisition cycles. Visualizations that illustrate the ACC/NLL progression for all datasets throughout these cycles can be found in Appendix C. REGrad* demonstrates superior performance over various baselines. For example, it shows a 1.72\% improvement in ACC over ExGrad. Compared to perturbation-based methods, REGrad* outperforms Perturb $x$ with a 2.36\% increase in ACC. Moreover, compared to LA, REGrad* achieves significant enhancements in both ACC and NLL. These improvements are particularly noteworthy given that training data is limited.

\subsection{Ablation Studies}
\textbf{Effectiveness of Component Contributions.} This section delves into the individual impact of REGrad, layer-selective gradients, and perturbations in OOD detection. The detailed results are presented in Table \ref{tab:each}. It shows that REGrad itself can achieve enhanced performance compared to ExGrad, with further improvements observed when layer-selective gradients are added. The combination of REGrad with both layer-selective gradients and perturbations yields the best results, indicating the cumulative positive impact of these components. More analysis is shown in Appendix D. 
\vspace{-3mm}

\begin{table}[h]\fontsize{7}{9}\selectfont
    \centering
    \vspace{-2mm}
	\caption{OOD detection for ablation studies using AUROC/AUPR.}
	\label{tab:each}
\begin{tabular}{|l|c|c|}
\hline
	Method  &MNIST $\rightarrow$ FMNIST& C10 $\rightarrow$ SVHN \\ 
\hline  
\multirow{1}{*}{ExGrad}   &98.11/97.98& 88.87/84.85 \\
\multirow{1}{*}{REGrad}   &98.42/98.45& 89.64/85.75\\
\multirow{1}{*}{REGrad + layer-selective}   & 98.51/98.55 &90.33/87.45 \\
\multirow{1}{*}{REGrad + layer-selective + perturb}   & \textbf{98.78}/\textbf{98.79} &\textbf{92.32}/\textbf{89.42}\\
\hline
\end{tabular}

\vspace{+1mm}
\begin{tabular}{|l|c|c|}
\hline
	Method  &SVHN $\rightarrow$ C10& C100 $\rightarrow$ SVHN \\ 
\hline  
\multirow{1}{*}{ExGrad}   &88.57/85.55& 79.99/72.68\\
\multirow{1}{*}{REGrad}   &90.22/88.16  &81.43/81.12\\
\multirow{1}{*}{REGrad + layer-selective}   &90.72/88.82&87.32/83.39 \\
\multirow{1}{*}{REGrad + layer-selective + perturb}   &\textbf{91.11}/\textbf{89.63}& \textbf{88.06}/\textbf{85.74}\\
\hline
\end{tabular}
\end{table}\vspace{-2mm}

\noindent \textbf{Hyperparameter Analysis.} We analyze the coefficient $\lambda$ in layer-selective gradients which adjusts weights towards deeper layers. As $\lambda \rightarrow \infty$, only the last layer's parameters influence gradient computation, whereas $\lambda \rightarrow 0$ treats all parameters equally. Our study in Table \ref{tab:hyper} shows that the OOD detection performance is relatively stable within specific $\lambda$ ranges. Additionally, we examine the impact of $\sigma$ in perturbations, selecting small values as per Proposition \ref{proposition1}, and find the results are similarly stable within certain $\sigma$ ranges. More analysis is shown in Appendix D.

\noindent \textbf{Efficiency Analysis.} While the complexity of the layer-selective gradients and the perturbation-integrated gradients are discussed in Section 4, we provide the empirical runtime in Appendix D.

\vspace{-2mm}
\begin{table}[h]\fontsize{7}{9}\selectfont
    \centering
    \vspace{-2mm}
	\caption{OOD detection for hyperparameter analysis using AUROC/AUPR. We use $\lambda =0.3, \sigma = 0.02$ for our method.}
	\label{tab:hyper}
\begin{tabular}{|l|c|}
\hline
	Method  & SVHN$\rightarrow$ C10 \\ 
\hline  
\multirow{1}{*}{REGrad + layer-selective ($\lambda \rightarrow 0$)} &90.22/88.16 \\
\multirow{1}{*}{REGrad + layer-selective ($\lambda \rightarrow \infty$)} &88.21/83.92 \\
\multirow{1}{*}{REGrad + layer-selective ($\lambda =0.25$)} & 90.65/88.72\\
\multirow{1}{*}{REGrad + layer-selective ($\lambda =0.3$)} & 90.72/88.82\\
\multirow{1}{*}{REGrad + layer-selective ($\lambda =0.35$)} & 90.80/88.82\\
\multirow{1}{*}{REGrad + layer-selective ($\lambda =0.3$) + perturb ($\sigma = .015$)} & 91.05/89.55\\
\multirow{1}{*}{REGrad + layer-selective ($\lambda =0.3$) + perturb ($\sigma = .02$)} & \textbf{91.11}/\textbf{89.63}\\
\multirow{1}{*}{REGrad + layer-selective ($\lambda =0.3$) + perturb ($\sigma = .025$)} & 90.98/89.41\\
\hline
\end{tabular}\vspace{-2mm}
\end{table}
\noindent \textbf{Varying Model Architecture.} While the main evaluation focuses on pre-trained CNN-based models, we also assess our method's effectiveness on vision transformers in Appendix D, where it continues to outperform other baselines.

\section{Conclusion}

In our study, we present a novel gradient-based method for epistemic UQ in pre-trained models, beginning with a theoretical analysis of gradient-based and perturbation-based methods' capabilities in capturing epistemic uncertainty. To improve current gradient-based methods, we introduce class-specific gradient weighting, layer-selective gradients, and gradient-perturbation integration. The proposed method does not require original training data or model modifications, ensuring broad applicability across any architecture and training technique. Our experiments across diverse scenarios, including out-of-distribution detection, uncertainty calibration, and active learning, demonstrate the superiority of our method over current state-of-the-art UQ methods for pre-trained models. 
\\\\
\noindent \textbf{Acknowledgement.} This work is supported in part by the National Science
Foundation award IIS 2236026.

\clearpage
\setcounter{page}{1}
\maketitlesupplementary
\appendix

\section{Proofs of Propositions}
In this section, we will provide the proofs for all the propositions shown in Section 3. 
\subsection{Proof of Proposition \ref{proposition1}}

\begin{proof}
This proof is based on the Bernstein-von Mises theorem \cite{kleijn2012bernstein,gelman2011induction}. Under mild regularity conditions, such as continuity of the likelihood function of $\theta$ and the maximum likelihood estimate $\theta^*$ not being on the parameter space boundary, the posterior distribution of $\theta$ converges in distribution to a multivariate Gaussian distribution as the sample size $|\mathcal{D}|\rightarrow \infty$. Specifically, we have:
\begin{equation}
    \label{Bernstein-von Mises equation}
    \begin{split}
        &\sup_{\theta} |p(\theta|\mathcal{D}) - \mathcal{N}(\theta;\theta^*, |\mathcal{D}|^{-1}I(\theta^*)^{-1}))| \rightarrow 0 \\
        & \text{where} \quad \theta^* = \arg\max_{\theta} \log p(\mathcal{D}|\theta)
    \end{split}
\end{equation}
where $I(\theta^*)$ is the Fisher information matrix at $\theta^*$. Assuming $\theta^*$ is bounded and the likelihood function at $\theta^*$ is differentiable and continuous, the Fisher information matrix is also bounded. This is because the bounded $\theta^*$ and the continuity of the likelihood function preclude unbounded derivatives. As $|\mathcal{D}|\rightarrow \infty$, we observe:
\begin{equation}
    \label{close}
    \rVert |\mathcal{D}|^{-1}I(\theta^*)^{-1} - \sigma^2 I \lVert \rightarrow 0
\end{equation}
for a sufficiently small $\sigma$ approaching 0. Therefore, we conclude:
\begin{equation}
    \label{finish_proposition1_proof}
    \sup_{\theta} |p(\theta|\mathcal{D}) - \mathcal{N}(\theta;\theta^*,\sigma^2 I)| \rightarrow 0 
\end{equation}
as $|\mathcal{D}|\rightarrow \infty$ and $\sigma \rightarrow 0$.
\end{proof}

\subsection{Proof of Proposition \ref{upper_bound}}
\begin{proof}
We first introduce Lemma \ref{l1}, which is central to our proof.
\begin{lemma} \label{l1}
    Denote $v(\theta) = -\frac{1}{|\mathcal{D}|}\log p(\theta) - \frac{1}{|\mathcal{D}|} \sum_{(x,y) \in \mathcal{D}} \log p(y|x,\theta)$ with $p(\theta)$ as a pre-defined prior distribution. Under the regularity constraints on $v(\theta)$ from Sec. 2.2 of \cite{katsevich2023tight}, the total variation distance $\text{D}_{\text{TV}}(p(\theta|\mathcal{D}), \mathcal{N}(\theta;\theta^*, -H^{-1}))$ between the posterior distribution $p(\theta|\mathcal{D})$ and its Laplacian approximation $\mathcal{N}(\theta;\theta^*, -H^{-1})$ fulfills
\begin{equation}
\begin{split}
    &|\text{D}_{\text{TV}}(p(\theta|\mathcal{D}), \mathcal{N}(\theta;\theta^*, -H^{-1})) - L| \\
    &\leq \scalemath{1.0}{c (c_3^2(v) +c_4(v)) \frac{d^2}{|\mathcal{D}|} }
\end{split}
\end{equation}
$H(\theta^*) = \nabla_{\theta^*}^2 \log p(\theta^*|\mathcal{D})$ represents Hessian matrix. $L \in [0,c_3(v) \frac{d}{\sqrt{|\mathcal{D}|}}]$ is an explicit function of $|\mathcal{D}|$. $d$ is the dimension of $\theta$, $c$ is an absolute constant, and $c_3(v), c_4(v)$ are constants computed from third/fourth-order derivatives of $v$.
\end{lemma}
The definition of $c, c_3(v), c_4(v)$ can be directly found in \cite{katsevich2023tight}. This lemma is directly obtained from  \cite{katsevich2023tight}. It is worth noting that $\mathcal{N}(\theta;\theta^*, -H^{-1})$ is the Laplacian approximation of $p(\theta|\mathcal{D})$. According to Sec. 3.3 of \cite{sharma2021sketching}, the Hessian matrix $-H$ can be well-approximated by the dataset fisher $|\mathcal{D}|F_{\theta^*}^{\mathcal{D}}$. For a detailed discussion, see \cite{sharma2021sketching}. This shows that $-H^{-1} \rightarrow \mathbf{0}$ when $|\mathcal{D}|\rightarrow \infty$. Lemma \ref{l1} further enables:
\begin{equation}
\begin{split}
    &\text{D}_{\text{TV}}(p(\theta|\mathcal{D}), \mathcal{N}(\theta;\theta^*, -H^{-1})) \\
    & \leq \scalemath{1.0}{c (c_3^2(v) +c_4(v)) \frac{d^2}{|\mathcal{D}|} } + |c_3(v)| \frac{d}{\sqrt{|\mathcal{D}|}}
\end{split}
\end{equation}

Then, based on Lemma \ref{l1}, we have
\begin{equation}
    \begin{split}
        &\text{D}_{\text{TV}}(p(\theta|\mathcal{D}), N(\theta;\theta^*,\sigma^2 I)) =\sup_{\theta} |p(\theta|\mathcal{D}) - \mathcal{N}(\theta;\theta^*, \sigma^2 I)| \\
        & \leq \sup_{\theta} \left |p(\theta|\mathcal{D}) -  \mathcal{N}(\theta^*, -H^{-1})\right| \\
        & + \sup_{\theta} \left|\mathcal{N}(\theta^*, -H^{-1}) - \mathcal{N}(\theta^*, \sigma^2 I)\right| \\
        & \leq \scalemath{1.0}{c (c_3^2(v) +c_4(v)) \frac{d^2}{|\mathcal{D}|} } + |c_3(v)| \frac{d}{\sqrt{|\mathcal{D}|}} (\text{Lemma \ref{l1}})\\
        & + \sqrt{\frac{1}{2} \text{KL}(\mathcal{N}(\theta^*,\sigma^2 I)||\mathcal{N}(\theta^*,-H^{-1})} (\text{Pinsker's inequality \cite{canonne2022short}})
    \end{split}
\end{equation}
\end{proof}
\subsection{Proof of Proposition \ref{proposition2}}
\begin{proof}
When perturbations $\Delta \theta$ and $\Delta x$ are both small, we can perform the first-order Taylor expansion as follows:
\begin{equation} \label{first-order_Taylor_f}
    \begin{split}
        &f(x,\theta + \Delta \theta) = f(x,\theta) + \left ( \frac{\partial f(x,\theta)}{\partial \theta}\right )^T \Delta \theta \\
        &f(x + \Delta x,\theta ) = f(x,\theta) + \left ( \frac{\partial f(x,\theta)}{\partial x}\right )^T \Delta x.        
    \end{split}
\end{equation}
It is worth noting that $f(x,\theta + \Delta \theta) =  f(x + \Delta x,\theta)$ requires 
\begin{equation} \label{Taylor_equal_f}
    \left ( \frac{\partial f(x,\theta)}{\partial \theta}\right )^T \Delta \theta = \left ( \frac{\partial f(x,\theta)}{\partial x}\right )^T \Delta x 
\end{equation}
which generally holds true when $\Delta \theta \rightarrow 0$ and $\Delta x \rightarrow 0$. Furthermore, for any small $\Delta \theta$ that is not near zero, there also likely exists a solution to Eq.~\eqref{Taylor_equal_f} when $\Delta x$ is treated as an unknown. However, this solution may not be unique. This situation arises because the matrix $\left ( \frac{\partial f(x,\theta)}{\partial x}\right )$ is often of full rank, given that the dimension of $f$ is usually much lower than that of $x$ and $\theta$. Additionally, the gradient $\frac{\partial f(x,\theta)}{\partial x}$ tends to be noisy within the context of neural networks.

Similarly, for any perturbation $\Delta x$, we can always find a corresponding $\Delta \theta$ to satisfy Eq.~\eqref{Taylor_equal_f}. More specifically, we only need to find a perturbation on the first layer parameters.

Let's denote $\theta = \{\theta^l\}$ where $\theta^l$ represents the parameters of the $l$th layer of the neural network. Without loss of generality, we can assume $\theta^1$ is linearly connected with $x$ without the bias variable, i.e.,
\begin{equation}
    h^1_j = \text{ReLU}((\theta^1)^T x) = \text{ReLU}(\sum_{k} \theta^1_{kj} x_k)
\end{equation}
where $h^1_j$ is the $j$th neuron in the first hidden layer. $x_k$ is the $k$th element of $x$ and $\theta^1_{kj}$ is the corresponding weight linearly connecting $x_k$ to $h^1_j$. Applying corresponding perturbations $\Delta \theta^1_{kj}$ and $\Delta x_{k}$ leads to
\begin{equation}\label{perturbations_equal_h1}
    \sum_{k} (\theta^1_{kj} +\Delta \theta^1_{kj})  x_k = \sum_{k} \theta^1_{kj}  (x_k+ \Delta x_k).
\end{equation}
The sufficient and necessary condition is 
\begin{equation}\label{nc_h1}
    \Delta \theta^1_{kj} x_k = \theta^1_{kj} \Delta x_k
\end{equation}
which shows that we can find the corresponding first-layer parameter perturbations to mimic the perturbations on the input space. 
\end{proof}
\label{sec:rationale}

\subsection{Proof of Proposition \ref{p3}}

\begin{proof}
\textbf{1. Global optimality of $\theta^*$}

Based on the Bernstein-von Mises theorem \cite{kleijn2012bernstein,gelman2011induction} and Proposition 1, we know that with a sufficiently large number of training samples, the posterior distribution of parameters approximates a Gaussian distribution. The mean of this Gaussian distribution is the maximum likelihood estimate $\theta^*$, and the covariance matrix approaches $\sigma^2 I$ with $\sigma$ sufficiently small and approaching 0. It's important to note that this proposition primarily focuses on the neighborhood of an in-distribution $x$ rather than the entire training set $\mathcal{D}$. In contrast, our focus is on an unlimited number of training samples in the neighborhood of $x$, denoted as $\mathcal{D}(x)$, which is well-represented by the classification model $\theta^*$. By applying Proposition 1 directly, we obtain:

\begin{equation}
\sup_{\theta} |p(\theta|\mathcal{D}(x)) - \mathcal{N}(\theta; \theta^*,\sigma^2 I)| \rightarrow 0
\end{equation}

where $ \theta^* = \arg\max_{\theta} \log p(\mathcal{D}(x)|\theta)$. As the posterior distribution $p(\theta|\mathcal{D}(x))$ is a Gaussian distribution with a single mode at $\theta^*$, it follows that $\theta^*$ is the global optimum.

\textbf{2. The gradients are close to 0}

Given $f(x,\theta^*) = \log p(y|x, \theta^*)$ for classification problems, we aim to prove:
\begin{equation}\label{gradients_0}
\begin{split}
    & \frac{\partial \log p(y|x, \theta^*)}{\partial \theta^*} = 0 \\
    & \frac{\partial \log p(y|x+\Delta x, \theta^*)}{\partial \theta^*} = 0 \quad\quad \forall x +\Delta x \in \mathcal{N}(x).
\end{split}
\end{equation} 
However, we only know that $\theta^*$ is the maximum likelihood estimate (MLE) given all the samples in the neighborhood of $x$. Since $\theta^*$ is the MLE, we have:
\begin{equation}
    \label{gradient_theta_D(x)}
    \frac{\partial \log p(D(x)|\theta^*)}{\partial \theta^*} =  \frac{\partial \sum_{\Delta x} \log p(y|x+\Delta x, \theta^*)}{\partial \theta^*} =0.
\end{equation}
We assume the neighborhood of $x$ is small ($\Delta x$ is small) and they share the same label $y$. Without loss of generality, we also consider that the neighborhood of $x$ is a ball centered at $x$ and $D(x)$ is sufficiently large to cover all possible $\Delta x$ in the ball. By performing the first-order Taylor expansion on $\log p(y|x+\Delta x, \theta^*)$ at $x$ when $\Delta x$ is small, we have
\begin{equation}
    \label{taylor_x}
    \begin{split}
         \log p(y|x+\Delta x, \theta^*) &= \log p(y|x,\theta^*) \\
         & + \left ( \frac{\partial \log p(y|x,\theta^*)}{\partial x}\right )^T \Delta x.
    \end{split}
\end{equation}
Combining Eq.~\eqref{gradient_theta_D(x)} and Eq.~\eqref{taylor_x}, we can derive that
\begin{equation}\label{gradients_x_to0}
    \begin{split}
        &\frac{\partial \sum_{\Delta x} \log p(y|x+\Delta x, \theta^*)}{\partial \theta^*} \\
        & = \sum_{\Delta x} \frac{\partial  \left (\log p(y|x,\theta^*) + \left ( \frac{\partial \log p(y|x,\theta^*) }{\partial x}\right )^T \Delta x \right )}{\partial \theta^*} \\
        & = \sum_{\Delta x} \frac{\partial  \log p(y|x, \theta^*)}{\partial \theta^*}+\sum_{\Delta x} \frac{\partial \log p(y|x,\theta^*) }{\partial x \partial \theta^*} \Delta x \\
        & = |D(x)| \frac{\partial  \log p(y|x, \theta^*)}{\partial \theta^*} \\
        &~~~~~~~+\frac{\partial \log p(y|x,\theta^*) }{\partial x \partial \theta^*} \left ( \sum_{\Delta x} \Delta x \right) \\  
        & \xrightarrow[]{|\mathcal{D}(x)|\rightarrow \infty} |D(x)| \frac{\partial  \log p(y|x, \theta^*)}{\partial \theta^*} = 0.
    \end{split}
\end{equation}
It is important to note that the matrix $\frac{\partial \log p(y|x,\theta^*) }{\partial x \partial \theta^*}$ has a dimension $|\theta| \times |x|$, and $\sum_{\Delta x} \Delta x = 0$ since positive and negative perturbations negate each other. Eq.~\eqref{gradients_x_to0} indicates that  
\begin{equation}
    \label{final_gradient_to0_0}
    \frac{\partial \log p(y|x,\theta^*)}{\partial \theta^*} = 0.
\end{equation}
Similarly, 
\begin{equation}\label{final_gradient_to0_1}
    \begin{split}
        &\frac{\partial \log p(y|x+\Delta x, \theta^*)}{\partial \theta^*} \\
        &= \frac{\partial \log p(y|x,\theta^*)}{\partial \theta^*} + \frac{\partial \log p(y|x,\theta^*) }{\partial x \partial \theta^*} \Delta x \\
        &=\frac{\partial \log p(y|x,\theta^*) }{\partial \theta^* \partial x} \Delta x = \frac{\partial \frac{\partial \log p(y|x,\theta^*) }{\partial \theta^*}}{\partial x} \Delta x\\
        &=\frac{\partial 0}{\partial x} \Delta x = 0.
    \end{split}
\end{equation}
This derivation relies on the assumption that the second partial derivatives of $\log p(y|x, \theta^*)$ are continuous, leading to the equality $\frac{\partial \log p(y|x,\theta^*) }{\partial \theta^* \partial x} = \frac{\partial \log p(y|x,\theta^*) }{\partial x \partial \theta^*}$ as per Clairaut's theorem. Clairaut's theorem states that for a function with continuous second partial derivatives, the differentiation order is immaterial.
\end{proof}

\subsection{Proof of Proposition \ref{p4}}
\begin{proof}
When $\Delta \theta \rightarrow 0$ and by using Taylor expansion for $\log p(y|x, \theta^*+\Delta \theta)$ at $\theta^*$, we can derive 
    \begin{equation} \small
    \begin{split}
        &\mathbb{E}_{p(\Delta \theta)} \left[ \text{KL}(p(y|x,\theta^*)||p(y|x,\theta^*+ \Delta \theta))\right] \\
        & = \mathbb{E}_{p(\Delta \theta)}\left[\sum_{c=1}^C p(y=c|x,\theta^*) \log \frac{p(y=c|x,\theta^*)}{p(y=c|x,\theta^*+\Delta \theta)}\right] \\
        & = \mathbb{E}_{p(\Delta \theta)}\left[-\sum_{c=1}^C p(y=c|x,\theta^*) \left ( \frac{\partial \log p(y=c|x,\theta^*)}{\partial \theta^*}\right)^T \Delta \theta \right] \\
        &=\mathbb{E}_{p(\Delta \theta)} \left [ \left \lVert -\sum_{c=1}^C p(y=c|x,\theta^*) \left ( \frac{\partial \log p(y=c|x,\theta^*)}{\partial \theta^*}\right)^T \Delta \theta \right\rVert \right ]\\
        &=\mathbb{E}_{p(\Delta \theta)} \left [\left \lVert \sum_{c=1}^C p(y=c|x,\theta^*) \left ( \frac{\partial \log p(y=c|x,\theta^*)}{\partial \theta^*}\right)^T \Delta \theta \right\rVert \right ]\\
        &\leq \mathbb{E}_{p(\Delta \theta)} \left [ \sum_{c=1}^C p(y=c|x,\theta^*) \left \lVert \left ( \frac{\partial \log p(y=c|x,\theta^*)}{\partial \theta^*}\right)^T \right\rVert \lVert \Delta \theta \rVert \right ] \\
        & = \sum_{c=1}^C p(y=c|x,\theta^*)\left \lVert \frac{\partial \log p(y=c|x,\theta^*)}{\partial \theta^*} \right \rVert \mathbb{E}_{p(\Delta \theta)} [||\Delta \theta||] \\
        & \propto \mathbb{E}_{y \sim p(y|x,\theta^*)}\left [ \left \lVert \frac{\partial \log p(y|x,\theta^*)}{\partial \theta^*} \right \rVert \right ] 
    \end{split}
\end{equation}
\end{proof}
\vspace{-3mm}
\section{Experiment Settings and Implementation Details}

\subsection{The Pre-trained Model}
While the pre-trained model could be more complex, we conduct experiments under the assumption of using a single deterministic neural network. Our goal is to demonstrate the effectiveness of our method in quantifying epistemic uncertainty, specifically within such a deterministic pre-trained network. We conducted experiments on four datasets: MNIST, C10, SVHN, and C100. For all training sessions, we randomly allocate 10\% of the training data as validation data for model selection. We utilize an RTX2080Ti GPU to perform all the experiments. Below, we detail the training procedures for the pre-trained models corresponding to each of the four datasets.

\begin{itemize}
    \item MNIST: We employ a simple CNN architecture: Conv2D-ReLU-Conv2D-ReLU-MaxPool2D-Dense-ReLU-Dense-Softmax. Each convolutional layer includes 32 filters with a $4\times 4$ kernel size. We utilize a max-pooling layer with a $2\times 2$ kernel and dense layers comprising 128 units. The SGD optimizer is used with a learning rate of 1e-2 and a momentum of 0.9. We set the maximum number of epochs at 30 and the weight decay coefficient at 5e-4. The batch size is 128.
    \item C10: We utilize ResNet18 for feature extraction, connected to a fully-connected layer for classification. The SGD optimizer is employed with an initial learning rate of 1e-1, decreasing to 1e-2, 1e-3, and 1e-4 at the 30th, 60th, and 90th epochs, respectively. The momentum is set at 0.9, with a maximum of 100 epochs and a weight decay coefficient of 5e-4. Standard data augmentation techniques, such as random cropping, horizontal flipping, and random rotation, are applied. The batch size is 128.
    \item  C100: ResNet152 is used for feature extraction, connected to a fully-connected layer for classification. The SGD optimizer starts with an initial learning rate of 1e-1, decreasing to 1e-2, 1e-3, and 1e-4 at the 40th, 70th, and 100th epochs, respectively. We keep the momentum at 0.9 and set the maximum number of epochs to 120, with a weight decay coefficient of 5e-4. Similar to C10, standard data augmentation techniques are employed. The batch size for C100 is 64.
    \item SVHN: We use a CNN architecture: Conv2D-ReLU-Conv2D-ReLU-MaxPool2D-Dense-ReLU-Dense-Dense-ReLU-Softmax. Each convolutional layer has 32 filters with a $4\times 4$ kernel size. A max-pooling layer with a $2\times 2$ kernel is used, along with dense layers having 128 units. The SGD optimizer starts with an initial learning rate of 0.05, reduced to 0.005, and 0.0005 at the 15th and 30th epochs. We set the maximum number of epochs at 50 and the weight decay coefficient at 5e-4. The batch size is 64.
\end{itemize}

\subsection{Implementation of Our Methods}

\begin{figure*}[h]
\centering
\includegraphics[width=12.2cm]{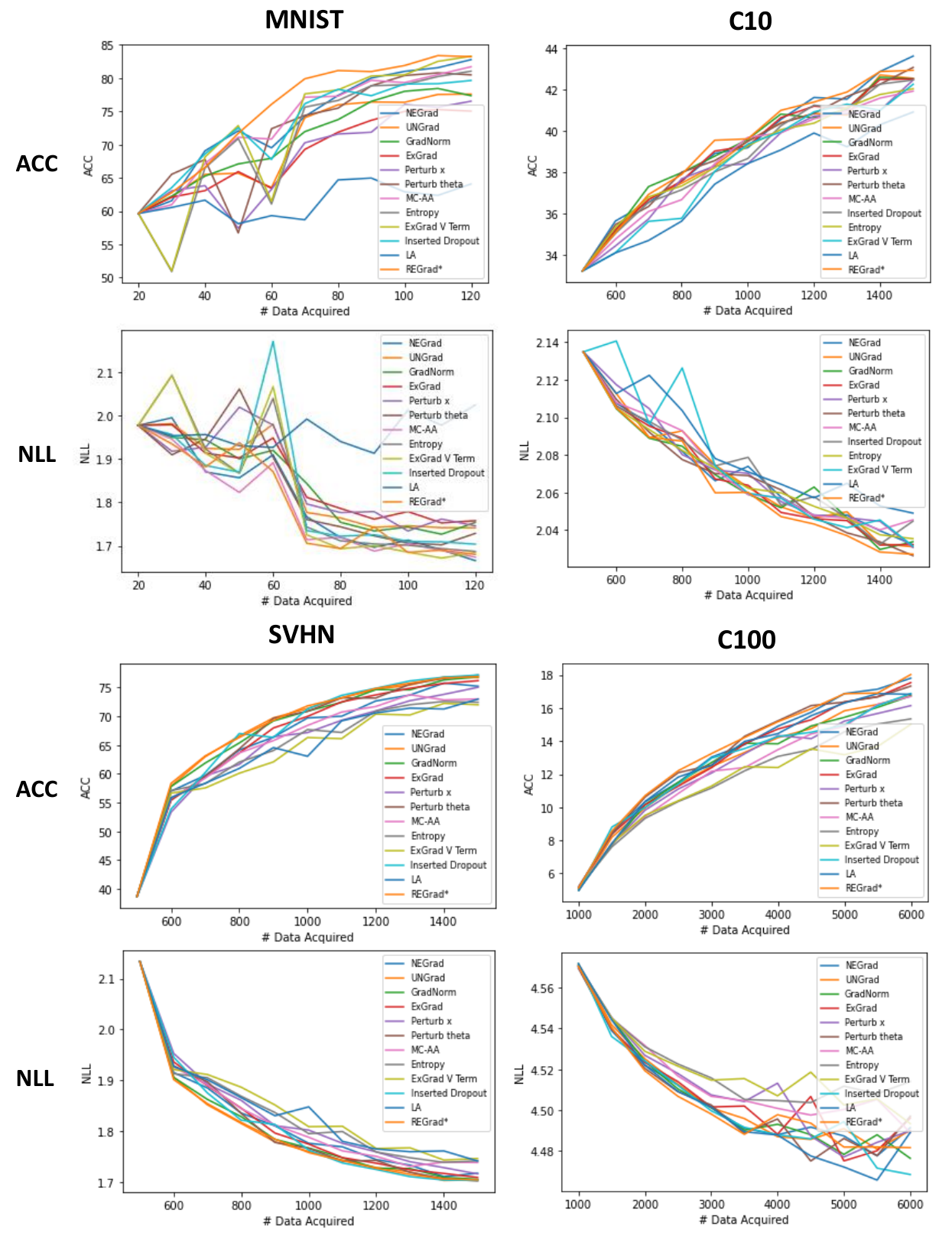}
\caption{ACC and NLL for the MNIST, C10, SVHN, and C100 datasets across 10 active learning acquisition cycles are presented. In each figure, the x-axis represents the number of data samples acquired for training the model, while the y-axis shows either the accuracy (ACC) or the negative log-likelihood (NLL) on the testing data. The results are averaged over three independent runs.}
\label{al_svhn_all}
\end{figure*}
\begin{itemize}
    \item Class-specific Gradient Weighting: We compute gradients using torch.autograd, which facilitates straightforward backpropagation from the model outputs to the model parameters.
    \item Layer-selective Gradients: The hyperparameter $\lambda$ was adjusted within the range of [0.1, 0.5], using increments of 0.05. Ultimately, the selected values of $\lambda$ were determined as 0.3 for the MNIST dataset, 0.2 for the C10 dataset, 0.2 for the SVHN dataset, and 0.4 for the C100 dataset.
    \item Perturbation-Gradient Integration: Following the guidelines of Proposition \ref{proposition1}, we selected a small value for the hyperparameter $\sigma$. It was tuned within the range [0.01, 0.05], in increments of 0.01. As a result, the value of $\sigma$ was set at 0.02 for all four datasets. Additionally, to generate perturbed inputs, we utilized 100 samples.
\end{itemize}

\subsection{Implementation of Baselines}
The mathematical formulation of all baseline methods is discussed in Section 5 of the main paper. Below, we provide detailed descriptions of their implementations.
\begin{itemize}
    \item NEGrad, UNGrad, GradNorm, ExGrad: These gradient-based methods are implemented using torch.autograd, and notably, they do not require any hyperparameter tuning. It's noteworthy that we experimented with both L1 and L2 norms for computing gradient norms, recording the best performing norm for each method.
    \item Perturb $x$: We introduce perturbations to the input $x$ by adding the Gaussian noise $\mathcal{N}(0,\sigma^2 I)$. The hyperparameter $\sigma$ is tuned in the range of [0.005,0.05]. Specifically, the values of $\sigma$ are determined to be 0.008 for the MNIST, C10, and SVHN datasets, and 0.02 for the C100 dataset. The number of perturbed inputs is set to 100.
    \item Perturb $\theta$: Perturbations are applied to all model parameters $\theta$ by adding the Gaussian noise from $\mathcal{N}(0,\sigma^2 I)$. The hyperparameter $\sigma$ is tuned within the range of [0.0001,0.01]. For the MNIST, C10, and SVHN datasets, the optimal $\sigma$ is found to be 0.008, while for C100 dataset, $\sigma$ is set to be 0.0003. The number of perturbed inputs is set to 100.
    \item MC-AA: The MC-AA approach involves introducing perturbations to the input using adversarial attacks. These attacks are performed through the Fast Gradient Sign Method (FGSM), as described in the following equation:
    \begin{equation}
    \label{FGSM}
    x_{\text{adv}} = x + \epsilon \cdot \text{sign}(\nabla_{x}L(\theta^*, x, y))
    \end{equation}
    In this equation, $L$ represents the negative log-likelihood loss, utilizing the predicted class as the label. The hyperparameter $\epsilon$ indicates the perturbation level, where sign($u$) is a function that outputs 1 if $u\geq0$ and $-1$ if $u<0$. To create multiple samples, we uniformly select $\epsilon$ from the range $[-a, a]$ with $a$ being tuned in the range of [0.0001, 0.01]. Based on this tuning process, we set $a$ at 0.0001 for MNIST, C10, and SVHN datasets, and at 0.0005 for the C100 dataset. The number of perturbations is set to 100.
    \item Inserted Dropout: In all pre-trained models, we incorporate an extra dropout layer just before the final fully connected layer. The dropout rate was subject to tuning, for which we explore values within the range [0.1, 0.5], using increments of 0.1. Ultimately, we set the dropout probability at 0.4 for all datasets.
    \item Entropy, ExGrad V Term: These can be directly calculated from the softmax output probabilities.
    \item LA: we employ the last-layer LA \cite{kristiadi2020being} with full Hessian matrix computation. We use the existing software proposed by \cite{daxberger2021laplace}, which is available at \url{https://github.com/AlexImmer/Laplace}. All hyperparameters are kept at their default settings as specified in the software.
\end{itemize}

\begin{table*}[ht]\small
\fontsize{8}{12}\selectfont
	\caption{The following two tables illustrate the effectiveness of component contributions for our method. The first table presents additional OOD detection results using AUROC (\%) $\uparrow$ and AUPR (\%) $\uparrow$. The second table displays uncertainty calibration results measured by rAULC. All experiments are aggregated over three independent runs.}
	\label{tab:ood_result_additional_ablation}
	\centering
\begin{tabular}{|l|c|c|c|c|c|l|}
\hline
	\multirow{2}{*}{Method} &\multicolumn{1}{|c|}{MNIST $\rightarrow$ Omniglot} &\multicolumn{1}{|c|}{C10 $\rightarrow$ LSUN} & \multicolumn{1}{|c|}{SVHN$ \rightarrow$ C100} & \multicolumn{1}{|c|}{C100$ \rightarrow$ LSUN}\\ \cline{2-5}
	&AUROC/AUPR &AUROC/AUPR  &AUROC/AUPR &AUROC/AUPR \\
\hline
 \multirow{1}{*}{ExGrad}   &97.55/96.99& 88.23/82.26 & 88.57/85.55 &76.93/71.00\\
\multirow{1}{*}{REGrad}   &97.80/97.54 & 90.15/87.16 &89.76/87.69 &78.95/73.71\\
\multirow{1}{*}{REGrad + layer-selective}   &98.01/97.86&90.79/88.86  &90.07/88.11 &79.01/\textbf{74.61}\\
\multirow{1}{*}{REGrad + layer-selective + perturb}   & \textbf{98.19}/\textbf{97.95} &\textbf{91.11}/\textbf{89.93} &\textbf{90.37}/\textbf{88.83} &\textbf{79.11}/74.39\\
\hline  
\end{tabular}

\vspace{+1mm}
\begin{tabular}{|l|c|c|c|c|c|l|}
\hline
	\multirow{2}{*}{Method} &\multicolumn{1}{|c|}{MNIST} &\multicolumn{1}{|c|}{C10} & \multicolumn{1}{|c|}{SVHN} & \multicolumn{1}{|c|}{C100}\\ \cline{2-5}
	&rAULC &rAULC  &rAULC &rAULC \\
\hline
 \multirow{1}{*}{ExGrad}   &0.985&0.893&0.868&0.841\\
\multirow{1}{*}{REGrad}   &0.985 &0.893&0.869 &0.846\\
\multirow{1}{*}{REGrad + layer-selective}  &0.985 &0.896 &0.870&0.854\\
\multirow{1}{*}{REGrad + layer-selective + perturb}  &\textbf{0.985}&\textbf{0.898}&\textbf{0.873}&\textbf{0.858} \\
\hline  
\end{tabular}
\end{table*}
\vspace{-2mm}
\begin{table*}[ht]\small
\fontsize{8}{12}\selectfont
	\caption{This table presents additional OOD detection results, highlighting the combination of layer-selective gradients and perturbation-integrated gradients with UNGrad and ExGrad. For UNGrad, the hyperparameters are set to $\lambda = 0.2, \sigma = 0.1$ for the SVHN dataset and $\lambda = 0.7, \sigma = 0.05$ for the C100 dataset. For ExGrad, $\lambda = 0.4, \sigma = 0.02$ are used for both the SVHN and C100 datasets. All experiments are aggregated over three independent runs.}
	\label{tab:ood_result_additional_ablation_other}
	\centering
\begin{tabular}{|l|c|c|c|c|c|l|}
\hline
	\multirow{2}{*}{Method} &\multicolumn{1}{|c|}{SVHN $\rightarrow$ C10} &\multicolumn{1}{|c|}{SVHN $\rightarrow$ C100} & \multicolumn{1}{|c|}{C100$ \rightarrow$ SVHN} & \multicolumn{1}{|c|}{C100$ \rightarrow$ LSUN}\\ \cline{2-5}
	&AUROC/AUPR &AUROC/AUPR  &AUROC/AUPR &AUROC/AUPR \\
\hline
 \multirow{1}{*}{UNGrad}   &68.28/69.21& 68.34/68.95 & 44.34/43.44 &47.46/45.41\\
\multirow{1}{*}{UNGrad + layer-selective} &75.41/75.52 &75.49/\textbf{75.11 }&82.44/75.87 &73.47/65.53\\
\multirow{1}{*}{UNGrad + layer-selective + perturb} &\textbf{78.71}/\textbf{76.54} &\textbf{77.48}/74.98 &\textbf{85.08}/\textbf{79.84} &\textbf{75.49}/\textbf{70.03}\\
 \multirow{1}{*}{ExGrad}   &88.57/85.55& 88.25/85.21 & 79.99/72.86 &76.93/71.00\\
\multirow{1}{*}{ExGrad + layer-selective}  &89.57/87.76 & 88.54/86.66 &80.91/73.90 &76.79/70.74\\
\multirow{1}{*}{ExGrad + layer-selective + perturb} &\textbf{90.57}/\textbf{89.64} &\textbf{89.64}/\textbf{88.58} &\textbf{81.09}/\textbf{74.39} &\textbf{78.23}/\textbf{73.43}\\
\hline  
\end{tabular}
\end{table*}

\subsection{Additional Information on Experiments}

\noindent \textbf{Uncertainty Calibration.} The implementation of metric rAULC is available at \url{https://github.com/janisgp/practicality_deterministic_epistemic_uncertainty}. 

\noindent \textbf{Active Learning.} 
\begin{itemize}
    \item Model Details: We utilize a simple CNN architecture for active learning across MNIST, C10, SVHN, and C100 datasets: Conv2D-ReLU-Conv2D-ReLU-MaxPool2D-Dense-ReLU-Dense-Softmax. Each convolutional layer includes 32 filters with a $4\times 4$ kernel size. We utilize a max-pooling layer with a $2\times 2$ kernel and dense layers comprising 128 units.
    \item Training Details:  We employ the SGD optimizer, configured with a learning rate of 0.01 and a momentum of 0.9. The weight decay is set at 0.0005, and we use a batch size of 128. In each active learning cycle, the model is trained for up to 200 epochs to ensure convergence. This training approach is consistently applied across all four datasets. Additionally, during each active learning cycle, the best-performing model on the validation data is saved.
    \item Splitting Training and Validation Data: Initially, we randomly select the first $m_1$ training samples from indices 1 to 10,000 in the training dataset. We ensure a balanced selection where each class is represented by $\frac{m_1}{C}$ samples. For validation purposes, we use the data samples indexed from 10,001 to 20,000. This validation set is crucial for model selection during the 200 training epochs of each active learning cycle. Samples indexed above 20,000 in the training dataset are reserved in an ``unused training pool", which serves as a source for acquiring new samples for subsequent retraining. In each cycle, we iteratively select $m_2$ samples from this pool based on the highest epistemic uncertainty. Upon completion of each training cycle, we assess the model's accuracy (ACC) and negative log-likelihood (NLL) using the original test dataset. We set $m_1 = 20, 500, 500, 1000$ and $m_2 = 20, 100, 100, 500$ for MNIST, C10, SVHN, and C100 datasets, respectively. 
    
\end{itemize}

\begin{table*}[h]\fontsize{8}{12}\selectfont
    \centering
    \vspace{-2mm}
	\caption{OOD detection for hyperparameter analysis is presented using AUROC(\%)/AUPR(\%). A ``***'' symbol indicates the hyperparameters used in the main paper.}
	\label{tab:hyper_additional}
\begin{tabular}{|l|c|c|c|}
\hline
	Method  & MNIST$\rightarrow$ Omniglot & MNIST$\rightarrow$ FMNIST\\ 
\hline  
\multirow{1}{*}{REGrad + layer-selective ($\lambda \rightarrow 0$)} &97.80/97.54 &98.42/98.45 \\
\multirow{1}{*}{REGrad + layer-selective ($\lambda \rightarrow \infty$)} &97.71/97.17 &97.70/97.27 \\
\multirow{1}{*}{REGrad + layer-selective ($\lambda =0.25$)} &98.01/97.86 & 98.38/98.40\\
\multirow{1}{*}{REGrad + layer-selective ($\lambda =0.3$)} &98.01/97.86 & 98.51/98.55\\
\multirow{1}{*}{REGrad + layer-selective ($\lambda =0.35$)} & 98.01/97.86 & 98.39/98.41\\
\multirow{1}{*}{REGrad + layer-selective ($\lambda =0.3$) + perturb ($\sigma = .015$)}  & 98.13/97.91 & 98.74/98.76 \\
\multirow{1}{*}{REGrad + layer-selective ($\lambda =0.3$) + perturb ($\sigma = .02$)***} & \textbf{98.19}/\textbf{97.95} & 98.78/98.79\\
\multirow{1}{*}{REGrad + layer-selective ($\lambda =0.3$) + perturb ($\sigma = .025$)} & 98.18/97.95 & \textbf{98.87}/\textbf{98.88}\\
\hline
\end{tabular}

\vspace{+1mm}
\begin{tabular}{|l|c|c|c|}
\hline
	Method  & C10$\rightarrow$ SVHN & C10$\rightarrow$ LSUN\\ 
\hline  
\multirow{1}{*}{REGrad + layer-selective ($\lambda \rightarrow 0$)} &89.64/85.75 &90.15/87.16 \\
\multirow{1}{*}{REGrad + layer-selective ($\lambda \rightarrow \infty$)} &89.87/86.76 &90.65/88.71 \\
\multirow{1}{*}{REGrad + layer-selective ($\lambda =0.15$)} &90.29/86.51 &90.94/88.11\\
\multirow{1}{*}{REGrad + layer-selective ($\lambda =0.2$)} &90.33/87.45 & 90.79/88.86\\
\multirow{1}{*}{REGrad + layer-selective ($\lambda =0.25$)} &90.45/86.92 &91.02/88.30\\
\multirow{1}{*}{REGrad + layer-selective ($\lambda =0.2$) + perturb ($\sigma = .01$)} &\textbf{92.50}/89.08 &91.01/89.03\\
\multirow{1}{*}{REGrad + layer-selective ($\lambda =0.2$) + perturb ($\sigma = .015$)}   &92.45/89.08 & 91.06/89.37\\
\multirow{1}{*}{REGrad + layer-selective ($\lambda =0.2$) + perturb ($\sigma = .02$)***} & 92.32/\textbf{89.42} & \textbf{91.11}/\textbf{89.93}\\
\hline
\end{tabular}

\vspace{+1mm}
\begin{tabular}{|l|c|c|c|}
\hline
	Method  & C100$\rightarrow$ SVHN & C100$\rightarrow$ LSUN\\ 
\hline  
\multirow{1}{*}{REGrad + layer-selective ($\lambda \rightarrow 0$)} &81.43/81.12 &78.95/73.71\\
\multirow{1}{*}{REGrad + layer-selective ($\lambda \rightarrow \infty$)}  &86.74/82.15 & 76.93/72.40 \\
\multirow{1}{*}{REGrad + layer-selective ($\lambda =0.35$)} &87.22/83.34 &78.66/74.12\\
\multirow{1}{*}{REGrad + layer-selective ($\lambda =0.4$)} &87.32/83.39 & 79.01/\textbf{74.61}\\
\multirow{1}{*}{REGrad + layer-selective ($\lambda =0.45$)} &87.38/83.25 &78.91/74.57\\
\multirow{1}{*}{REGrad + layer-selective ($\lambda =0.4$) + perturb ($\sigma = .015$)} &87.89/84.93 &78.98/74.45\\
\multirow{1}{*}{REGrad + layer-selective ($\lambda =0.4$) + perturb ($\sigma = .02$)***}   &\textbf{88.06}/\textbf{85.74} &\textbf{79.11}/74.39\\
\multirow{1}{*}{REGrad + layer-selective ($\lambda =0.4$) + perturb ($\sigma = .025$)} &88.01/85.52 &78.91/74.23\\
\hline
\end{tabular}

\end{table*}

\section{Additional Active Learning Results}
This section presents the visualizations of ACC and NLL progression for all four datasets shown in Figure \ref{al_svhn_all}. We arrive at the same conclusion as indicated in Table \ref{tab:active_learning}: our proposed method consistently surpasses various baselines in achieving the highest accuracy and the lowest negative log-likelihood. For the MNIST dataset, the noticeable variance in the lines is likely attributed to the process starting with only 20 samples and acquiring just 10 more samples each time. This large variance could be due to the randomness inherent in the limited sample size and the training process itself. For averaged ACCs and NLLs across the four datasets, please refer to Table \ref{tab:active_learning} in the main paper.

\section{Additional Ablation Studies}

\subsection{Effectiveness of Component Contributions}

Besides Table \ref{tab:each}, we also provide additional results for OOD detection and uncertainty calibration to demonstrate the effectiveness of each component. These results, detailed in Tables \ref{tab:ood_result_additional_ablation}, support similar conclusions to those in Section 5.4 of the main paper. Each component independently enhances performance, with the combination of all components yielding the best results. Although all components significantly improve OOD detection, the impact on uncertainty calibration becomes marginal. This may be partly due to the fact that a well-calibrated pre-trained model typically outputs predictive probabilities that closely match actual performance. By incorporating probability information through class-specific gradient weighting, our gradient-based method achieves good calibration performance, even without employing layer-selective and perturbation-integrated gradient strategies. This scenario presents a challenge in further enhancing calibration, particularly when focusing on pre-trained models. However, as model complexity increases, these models tend to be less calibrated. In such instances, employing layer-selective and perturbation-integrated gradient strategies becomes more advantageous for significant improvements. This is especially evident in the case of the C100 dataset.

We also demonstrate the effectiveness of layer-selective gradients and perturbation-integrated gradients when combined with other gradient-based methods such as ExGrad and UNGrad. The OOD detection results for the MNIST and C10 datasets are presented in Table \ref{tab:ood_result_additional_ablation_other}. From these results, we can conclude that the two techniques we propose can effectively enhance the performance of other gradient-based methods.

\subsection{Hyperparameter Analysis.}
In addition to the analysis presented in Table \ref{tab:hyper}, we further evaluate the impact of the coefficient $\lambda$ in the layer-selective gradients and $\sigma$ in the perturbation-integrated gradients for the MNIST, C10, and C100 datasets. These additional results are shown in Table \ref{tab:hyper_additional}, where we observe similar conclusions: the performance is not overly sensitive to the hyperparameters within certain ranges. It's also important to note that these hyperparameters were not determined through an exhaustive brute-force search of all possible parameter combinations. Furthermore, they were chosen based on their overall performance across various experimental tasks. While it's possible that some hyperparameters might perform better in specific cases, the current settings already provide the best performance in comparison to various baselines.

\begin{table}[h]\fontsize{8}{12}\selectfont
    \centering
    \vspace{-2mm}
	\caption{Runtime analysis for UQ in seconds (s).}
	\label{tab:efficency}
\begin{tabular}{|l|c|c|}
\hline
	Method  & C10 UQ Runtime \\ 
\hline  
\multirow{1}{*}{NEGrad } & 18.73s\\
\multirow{1}{*}{UNGrad } & 33.45s\\
\multirow{1}{*}{GradNorm} & 18.54s\\
\multirow{1}{*}{ExGrad } & 34.72s\\
\multirow{1}{*}{Perturb $x$} & 32.56s\\
\multirow{1}{*}{Perturb $\theta$} & 43.16s\\
\multirow{1}{*}{MC-AA} & 32.56s\\
\multirow{1}{*}{Inserted Dropout} & 16.15s\\
\multirow{1}{*}{Entropy} & 3.05s\\
\multirow{1}{*}{ExGrad V Term} & 2.52s\\
\multirow{1}{*}{LA} & 22.61s\\
\multirow{1}{*}{REGrad } & 33.28s\\
\multirow{1}{*}{REGrad + layer-selective} & 29.27s\\
\multirow{1}{*}{REGrad + layer-selective + perturb} & 75.87s\\
\hline
\end{tabular}
\end{table}
\vspace{-3mm}
\begin{table*}[h] 
    \centering
    \fontsize{8}{12}\selectfont
    \caption{ViT Results with AUROC(\%)/AUPR(\%).}
    \label{tab:vit}
\begin{tabular}{l|c c c c}
   \hline
   Method  &C10 $\rightarrow$ SVHN &C10 $\rightarrow$ LSUN &C100 $\rightarrow$ SVHN  &C100 $\rightarrow$ LSUN\\
   \hline
   UNGrad  & 78.77/72.51 &84.83/76.61&70.20/60.06&72.08/60.19\\
    GradNorm & 53.03/58.91 & 72.84/71.04 &56.30/54.84&63.76/57.20\\
    ExGrad &92.41/88.06 &89.06/86.55 &85.47/78.91&85.55/79.68\\ 
    Perturb $\theta$  &89.13/81.26&90.70/81.59 &84.35/77.37&85.70/77.74\\
    Ours &\textbf{96.62/95.69}&\textbf{94.99/94.06}&\textbf{90.22/88.61}&\textbf{88.38/85.10} \\
   \hline

\end{tabular}

\end{table*} 
\subsection{Efficiency Analysis for UQ.} While the complexity of both layer-selective gradients and perturbation-integrated gradients is discussed in Section 4, Table \ref{tab:efficency} provides their empirical runtime, specifically for computing the uncertainty of 1000 C10 images. Entropy-based methods are the most efficient, requiring only a single forward pass for uncertainty calculations. In comparison, gradient-based and perturbation-based methods demonstrate similar efficiency levels. Perturbation-based methods necessitate multiple forward passes, whereas gradient-based methods require both forward and backward passes. However, the efficiency of gradient-based methods can be enhanced through parallel computing. Due to the limitation of torch.autograd, which can only compute gradients of scalar functions of the input, we compute the gradients for each test sample sequentially (with a batch size of 1). For other baselines, except for gradient-based methods, we use a batch size of 32. Methods like ExGrad, UNGrad, and REGrad are more time-consuming than NEGrad and GradNorm because they require sequential computation of gradients for each class's probability. Although certain computations in class-wise gradient calculations can be reused, torch.autograd does not support parallel computation of gradients. Moreover, as discussed in Section 4 of the main paper, employing layer-wise gradient strategies does not add to the complexity. However, the perturbation-integrated gradient strategy requires additional time, mainly due to the need for extra forward propagation compared to other gradient-based methods.

\subsection{Varying Model Architecture} We further train the model using a Vision Transformer (ViT) for the C10 and C100 datasets. The training procedure adheres to the guidelines provided in the GitHub repository at \url{https://github.com/kentaroy47/vision-transformers-cifar10}. We utilize the ViT-timm model, following the default hyperparameters. As shown in Table \ref{tab:vit}, the OOD detection results for the C10 and C100 datasets demonstrate improved performance using ViT compared to ResNet. Our method continues to outperform baseline approaches.


\begin{thebibliography}{41}
\providecommand{\natexlab}[1]{#1}
\providecommand{\url}[1]{\texttt{#1}}
\expandafter\ifx\csname urlstyle\endcsname\relax
  \providecommand{\doi}[1]{doi: #1}\else
  \providecommand{\doi}{doi: \begingroup \urlstyle{rm}\Url}\fi

\bibitem[Alarab and Prakoonwit(2022)]{alarab2022adversarial}
Ismail Alarab and Simant Prakoonwit.
\newblock Adversarial attack for uncertainty estimation: identifying critical regions in neural networks.
\newblock \emph{Neural Processing Letters}, 54\penalty0 (3):\penalty0 1805--1821, 2022.

\bibitem[Alarab and Prakoonwit(2023)]{alarab2023uncertainty}
Ismail Alarab and Simant Prakoonwit.
\newblock Uncertainty estimation based adversarial attack in multi-class classification.
\newblock \emph{Multimedia Tools and Applications}, 82\penalty0 (1):\penalty0 1519--1536, 2023.

\bibitem[Canonne(2022)]{canonne2022short}
Cl{\'e}ment~L Canonne.
\newblock A short note on an inequality between kl and tv.
\newblock \emph{arXiv preprint arXiv:2202.07198}, 2022.

\bibitem[Chen et~al.(2014)Chen, Fox, and Guestrin]{SGHMC_chen2014}
Tianqi Chen, Emily~B. Fox, and Carlos Guestrin.
\newblock Stochastic gradient hamiltonian monte carlo.
\newblock In \emph{Proceedings of the 31st International Conference on International Conference on Machine Learning - Volume 32}, page II–1683–II–1691. JMLR.org, 2014.

\bibitem[Daxberger et~al.(2021{\natexlab{a}})Daxberger, Kristiadi, Immer, Eschenhagen, Bauer, and Hennig]{daxberger2021laplace}
Erik Daxberger, Agustinus Kristiadi, Alexander Immer, Runa Eschenhagen, Matthias Bauer, and Philipp Hennig.
\newblock Laplace redux-effortless bayesian deep learning.
\newblock \emph{Advances in Neural Information Processing Systems}, 34, 2021{\natexlab{a}}.

\bibitem[Daxberger et~al.(2021{\natexlab{b}})Daxberger, Nalisnick, Allingham, Antor{\'a}n, and Hern{\'a}ndez-Lobato]{daxberger2021bayesian}
Erik Daxberger, Eric Nalisnick, James~U Allingham, Javier Antor{\'a}n, and Jos{\'e}~Miguel Hern{\'a}ndez-Lobato.
\newblock Bayesian deep learning via subnetwork inference.
\newblock In \emph{International Conference on Machine Learning}, pages 2510--2521. PMLR, 2021{\natexlab{b}}.

\bibitem[Deng(2012)]{deng2012mnist}
Li Deng.
\newblock The mnist database of handwritten digit images for machine learning research.
\newblock \emph{IEEE Signal Processing Magazine}, 29\penalty0 (6):\penalty0 141--142, 2012.

\bibitem[Franchi et~al.(2020)Franchi, Bursuc, Aldea, Dubuisson, and Bloch]{franchi2020encoding}
Gianni Franchi, Andrei Bursuc, Emanuel Aldea, S{\'e}verine Dubuisson, and Isabelle Bloch.
\newblock Encoding the latent posterior of bayesian neural networks for uncertainty quantification.
\newblock \emph{arXiv preprint arXiv:2012.02818}, 2020.

\bibitem[Gal and Ghahramani(2016)]{gal2016dropout}
Yarin Gal and Zoubin Ghahramani.
\newblock Dropout as a bayesian approximation: Representing model uncertainty in deep learning.
\newblock In \emph{international conference on machine learning}, pages 1050--1059. PMLR, 2016.

\bibitem[Gelman(2011)]{gelman2011induction}
Andrew Gelman.
\newblock Induction and deduction in bayesian data analysis.
\newblock 2011.

\bibitem[Geman and Geman(1984)]{geman1984stochastic}
Stuart Geman and Donald Geman.
\newblock Stochastic relaxation, gibbs distributions, and the bayesian restoration of images.
\newblock \emph{IEEE Transactions on pattern analysis and machine intelligence}, \penalty0 (6):\penalty0 721--741, 1984.

\bibitem[Huang et~al.(2021)Huang, Geng, and Li]{huang2021importance}
Rui Huang, Andrew Geng, and Yixuan Li.
\newblock On the importance of gradients for detecting distributional shifts in the wild.
\newblock \emph{Advances in Neural Information Processing Systems}, 34:\penalty0 677--689, 2021.

\bibitem[Igoe et~al.(2022)Igoe, Chung, Char, and Schneider]{igoe2022useful}
Conor Igoe, Youngseog Chung, Ian Char, and Jeff Schneider.
\newblock How useful are gradients for ood detection really?
\newblock \emph{arXiv preprint arXiv:2205.10439}, 2022.

\bibitem[Katsevich(2023)]{katsevich2023tight}
Anya Katsevich.
\newblock Tight dimension dependence of the laplace approximation.
\newblock \emph{arXiv preprint arXiv:2305.17604}, 2023.

\bibitem[Kleijn and van~der Vaart(2012)]{kleijn2012bernstein}
Bas~JK Kleijn and Aad~W van~der Vaart.
\newblock The bernstein-von-mises theorem under misspecification.
\newblock \emph{Electronic Journal of Statistics}, 6:\penalty0 354--381, 2012.

\bibitem[Kristiadi et~al.(2020)Kristiadi, Hein, and Hennig]{kristiadi2020being}
Agustinus Kristiadi, Matthias Hein, and Philipp Hennig.
\newblock Being bayesian, even just a bit, fixes overconfidence in relu networks.
\newblock In \emph{International Conference on Machine Learning}, pages 5436--5446. PMLR, 2020.

\bibitem[Krizhevsky et~al.(2009)Krizhevsky, Hinton, et~al.]{krizhevsky2009learning}
Alex Krizhevsky, Geoffrey Hinton, et~al.
\newblock Learning multiple layers of features from tiny images.
\newblock 2009.

\bibitem[Krizhevsky et~al.(2014)Krizhevsky, Nair, and Hinton]{krizhevsky2014cifar}
Alex Krizhevsky, Vinod Nair, and Geoffrey Hinton.
\newblock The cifar-10 dataset.
\newblock \emph{online: http://www. cs. toronto. edu/kriz/cifar. html}, 55\penalty0 (5), 2014.

\bibitem[Lake et~al.(2015)Lake, Salakhutdinov, and Tenenbaum]{lake2015human}
Brenden~M Lake, Ruslan Salakhutdinov, and Joshua~B Tenenbaum.
\newblock Human-level concept learning through probabilistic program induction.
\newblock \emph{Science}, 350\penalty0 (6266):\penalty0 1332--1338, 2015.

\bibitem[Ledda et~al.(2023)Ledda, Fumera, and Roli]{ledda2023dropout}
Emanuele Ledda, Giorgio Fumera, and Fabio Roli.
\newblock Dropout injection at test time for post hoc uncertainty quantification in neural networks.
\newblock \emph{arXiv preprint arXiv:2302.02924}, 2023.

\bibitem[Lee and AlRegib(2020)]{lee2020gradients}
Jinsol Lee and Ghassan AlRegib.
\newblock Gradients as a measure of uncertainty in neural networks.
\newblock In \emph{2020 IEEE International Conference on Image Processing (ICIP)}, pages 2416--2420. IEEE, 2020.

\bibitem[Lee et~al.(2020)Lee, Humt, Feng, and Triebel]{lee2020estimating}
Jongseok Lee, Matthias Humt, Jianxiang Feng, and Rudolph Triebel.
\newblock Estimating model uncertainty of neural networks in sparse information form.
\newblock In \emph{International Conference on Machine Learning}, pages 5702--5713. PMLR, 2020.

\bibitem[Louizos and Welling(2017)]{louizos2017multiplicative}
Christos Louizos and Max Welling.
\newblock Multiplicative normalizing flows for variational bayesian neural networks.
\newblock In \emph{International Conference on Machine Learning}, pages 2218--2227. PMLR, 2017.

\bibitem[MacKay(1992)]{MacKay1992}
David J.~C. MacKay.
\newblock {A Practical Bayesian Framework for Backpropagation Networks}.
\newblock \emph{Neural Computation}, 4\penalty0 (3):\penalty0 448--472, 1992.

\bibitem[Maddox et~al.(2019)Maddox, Izmailov, Garipov, Vetrov, and Wilson]{Maddox_NIPS19_SWAG}
Wesley~J Maddox, Pavel Izmailov, Timur Garipov, Dmitry~P Vetrov, and Andrew~Gordon Wilson.
\newblock A simple baseline for bayesian uncertainty in deep learning.
\newblock In \emph{Advances in Neural Information Processing Systems 32}, pages 13132--13143. Curran Associates, Inc., 2019.

\bibitem[Malinin and Gales(2018)]{Malinin_prior_network_NIPS18}
Andrey Malinin and Mark Gales.
\newblock Predictive uncertainty estimation via prior networks.
\newblock In \emph{Advances in Neural Information Processing Systems 31}, pages 7047--7058. Curran Associates, Inc., 2018.

\bibitem[Malinin et~al.(2019)Malinin, Mlodozeniec, and Gales]{malinin2019ensemble}
Andrey Malinin, Bruno Mlodozeniec, and Mark Gales.
\newblock Ensemble distribution distillation.
\newblock \emph{arXiv preprint arXiv:1905.00076}, 2019.

\bibitem[Mi et~al.(2022)Mi, Wang, Tian, He, and Shavit]{mi2022training}
Lu Mi, Hao Wang, Yonglong Tian, Hao He, and Nir~N Shavit.
\newblock Training-free uncertainty estimation for dense regression: Sensitivity as a surrogate.
\newblock In \emph{Proceedings of the AAAI Conference on Artificial Intelligence}, pages 10042--10050, 2022.

\bibitem[Netzer et~al.(2011)Netzer, Wang, Coates, Bissacco, Wu, and Ng]{netzer2011reading}
Yuval Netzer, Tao Wang, Adam Coates, Alessandro Bissacco, Bo Wu, and Andrew~Y Ng.
\newblock Reading digits in natural images with unsupervised feature learning.
\newblock 2011.

\bibitem[Postels et~al.(2021)Postels, Segu, Sun, Van~Gool, Yu, and Tombari]{postels2021practicality}
Janis Postels, Mattia Segu, Tao Sun, Luc Van~Gool, Fisher Yu, and Federico Tombari.
\newblock On the practicality of deterministic epistemic uncertainty.
\newblock \emph{arXiv preprint arXiv:2107.00649}, 2021.

\bibitem[Ritter et~al.(2018)Ritter, Botev, and Barber]{Ritter_ICLR18_laplace}
Hippolyt Ritter, Aleksandar Botev, and David Barber.
\newblock {A scalable laplace approximation for neural networks}.
\newblock In \emph{6th International Conference on Learning Representations, ICLR 2018 - Conference Track Proceedings}, 2018.

\bibitem[Robert et~al.(1999)Robert, Casella, and Casella]{robert1999monte}
Christian~P Robert, George Casella, and George Casella.
\newblock \emph{Monte Carlo statistical methods}.
\newblock Springer, 1999.

\bibitem[Schweighofer et~al.(2023)Schweighofer, Aichberger, Ielanskyi, Klambauer, and Hochreiter]{schweighofer2023quantification}
Kajetan Schweighofer, Lukas Aichberger, Mykyta Ielanskyi, G{\"u}nter Klambauer, and Sepp Hochreiter.
\newblock Quantification of uncertainty with adversarial models.
\newblock \emph{arXiv preprint arXiv:2307.03217}, 2023.

\bibitem[Sharma et~al.(2021)Sharma, Azizan, and Pavone]{sharma2021sketching}
Apoorva Sharma, Navid Azizan, and Marco Pavone.
\newblock Sketching curvature for efficient out-of-distribution detection for deep neural networks.
\newblock In \emph{Uncertainty in artificial intelligence}, pages 1958--1967. PMLR, 2021.

\bibitem[Smilkov et~al.(2017)Smilkov, Thorat, Kim, Vi{\'e}gas, and Wattenberg]{smilkov2017smoothgrad}
Daniel Smilkov, Nikhil Thorat, Been Kim, Fernanda Vi{\'e}gas, and Martin Wattenberg.
\newblock Smoothgrad: removing noise by adding noise.
\newblock \emph{arXiv preprint arXiv:1706.03825}, 2017.

\bibitem[Teye et~al.(2018)Teye, Azizpour, and Smith]{teye2018bayesian}
Mattias Teye, Hossein Azizpour, and Kevin Smith.
\newblock Bayesian uncertainty estimation for batch normalized deep networks.
\newblock In \emph{International Conference on Machine Learning}, pages 4907--4916. PMLR, 2018.

\bibitem[Tierney(1994)]{tierney1994markov}
Luke Tierney.
\newblock Markov chains for exploring posterior distributions.
\newblock \emph{the Annals of Statistics}, pages 1701--1728, 1994.

\bibitem[Welling and Teh(2011)]{welling2011bayesian}
Max Welling and Yee~W Teh.
\newblock Bayesian learning via stochastic gradient langevin dynamics.
\newblock In \emph{Proceedings of the 28th international conference on machine learning (ICML-11)}, pages 681--688, 2011.

\bibitem[Xiao et~al.(2017)Xiao, Rasul, and Vollgraf]{xiao2017fashion}
Han Xiao, Kashif Rasul, and Roland Vollgraf.
\newblock Fashion-mnist: a novel image dataset for benchmarking machine learning algorithms.
\newblock \emph{arXiv preprint arXiv:1708.07747}, 2017.

\bibitem[Yu et~al.(2015)Yu, Zhang, Song, Seff, and Xiao]{journals/corr/YuZSSX15}
Fisher Yu, Yinda Zhang, Shuran Song, Ari Seff, and Jianxiong Xiao.
\newblock Lsun: Construction of a large-scale image dataset using deep learning with humans in the loop.
\newblock \emph{CoRR}, abs/1506.03365, 2015.

\bibitem[Zhang et~al.(2019)Zhang, Li, Zhang, Chen, and Wilson]{zhang2019cyclical}
Ruqi Zhang, Chunyuan Li, Jianyi Zhang, Changyou Chen, and Andrew~Gordon Wilson.
\newblock Cyclical stochastic gradient mcmc for bayesian deep learning.
\newblock \emph{arXiv preprint arXiv:1902.03932}, 2019.

\end{thebibliography}
\end{document}